\documentclass[a4paper]{article}
\usepackage[utf8]{inputenc}
\usepackage{geometry}
\usepackage{microtype}
\usepackage{graphicx}
\usepackage{subfigure}
\usepackage{booktabs} % for professional tables
\usepackage{microtype}
\usepackage{multirow}
\usepackage{framed}
\usepackage{graphicx,stfloats}
\usepackage{subfigure}
\usepackage{booktabs} % for professional tables
\usepackage{amsmath,amssymb,amsfonts}
\allowdisplaybreaks
\usepackage{mathtools}
\usepackage{amsthm}
\allowdisplaybreaks[4]
\usepackage{float,bbm}
\usepackage{titling}
\usepackage{setspace}
\usepackage{algorithm,algorithmic}


\usepackage{hyperref}
\usepackage[capitalize,noabbrev]{cleveref}

\theoremstyle{theorem}
\newtheorem{theorem}{Theorem}[section]

\newtheorem{lemma}{Lemma}[section]

\theoremstyle{definition}

\theoremstyle{remark}
\newtheorem{remark}{Remark}[section]

\usepackage[textsize=tiny]{todonotes}
\hypersetup{hidelinks}

\renewenvironment{framed}[1][\hsize]
{\MakeFramed{\hsize#1\advance\hsize-\width \FrameRestore}}%
{\endMakeFramed}

\begin{document}

\begin{center}
	{\Large Asymptotic Optimality of Myopic Ranking and Selection Procedures}
	\\\vspace{20pt}
	% Author names and affiliations
	~~Yanwen Li$^1$~~~~~~~~~~~~~~~~~~~Siyang Gao$^{2,3}$~~~~~~~~~~~~~~~Zhongshun Shi$^{4}$\\
	{\small \url{yanwen.li@hotmail.com}}~~~~{\small \url{siyangao@cityu.edu.hk}}~~~~{\small \url{tzshi@utk.edu}}
	\\\hspace{10pt}
	
	\small  
	$^1$ School of Physical and Mathematical Sciences, Nanyang Technological University, Singapore\\
	$^2$ Department of Advanced Design and Systems Engineering, City University of Hong Kong, Hong Kong\\
	$^3$ School of Data Science, City University of Hong Kong, Hong Kong\\
	$^4$ Department of Industrial and Systems Engineering, University of Tennessee-Knoxville, Knoxville, USA
\end{center}

\vspace{10pt}

\begin{abstract}
	Ranking and selection (R\&S) is a popular model for studying discrete-event dynamic systems. It aims to select the best design (the design with the largest mean performance) from a finite set, where the mean of each design is unknown and has to be learned by samples. Great research efforts have been devoted to this problem in the literature for developing procedures with superior empirical performance and showing their optimality. In these efforts, myopic procedures were popular. They select the best design using a ``naive'' mechanism of iteratively and myopically improving an approximation of the objective measure. Although they are based on simple heuristics and lack theoretical support, they turned out highly effective, and often achieved competitive empirical performance compared to procedures that were proposed later and shown to be asymptotically optimal. In this paper, we theoretically analyze these myopic procedures and prove that they also satisfy the optimality conditions of R\&S, just like some other popular R\&S methods. It explains the good performance of myopic procedures in various numerical tests, and provides good insight into the structure and theoretical development of efficient R\&S procedures.
\end{abstract}

\vspace{5pt}

\section{Introduction}\label{sec1}

Discrete-event dynamic systems (DEDS, \cite{ho1992}) are commonly seen in practice, such as in manufacturing plants, supply chains, electric power grids, traffic control systems, communication networks, etc. Since DEDS are often highly complicated, a primary and powerful tool for modeling and studying them is simulation \cite{chen2014b}. However, when running simulation experiments, the time cost could be substantial. For example, it may take several days to run a 3D extrusion simulation with the Finite Element Methodology in the turbine blade design problems \cite{ho2007}. It explains the broad research interests in improving the simulation efficiency in performance evaluation and optimization of DEDS.

The ranking and selection (R\&S) problem has been a prevailing model for optimizing DEDS and improving simulation efficiency \cite{chen2000a,chen2013}. It considers a finite number of designs (or decisions, solutions, alternatives, etc., under different problem contexts) with unknown means that can be learned by samples. The goal is to help decision-makers select the unique best design, defined as the one with the largest mean. Suppose we are given a limited sampling budget. We need to carefully determine the number of samples allocated to each design in order to produce better mean estimates for comparison. This is a general and useful model, and has attracted research interests in a variety of applications including manufacturing operations \cite{lee1997}, traffic management \cite{persula1999}, network reliability \cite{kiekhaefer2011}, etc. In some literature, it is also called ordinal optimization \cite{ho1992}.

The main methodological challenge in R\&S is the ``exploration-exploitation'' trade-off. We need to make a good balance between sampling designs that we do not know much about and sampling designs that appear to be good. The key question of interest is how R$\&$S procedures should allocate the sampling budget to the designs such that the true best design can be correctly selected.

When the unknown means are estimated by samples, it is impossible to correctly select the best design with probability one under a finite sampling budget. Therefore, it is natural to optimize some measures that can reasonably reflect the evidence of correct/incorrect selection. One such measure that has been intensively studied in R\&S is the probability of correct selection (PCS, \cite{chen2000a}) for measuring the evidence of correct selection. It refers to the probability that the estimated best design is identical to the true best one. Another is the expected opportunity cost (EOC, \cite{chick2001,gao2017}) for measuring the evidence of incorrect selection, where opportunity cost is defined as the difference between the means of the selected design and the true best design. EOC is the expectation of the opportunity cost. It incorporates both the chance of selecting the best design and the consequence of selecting a non-best one.

There are two popular classes of methods for solving the R\&S problem in the literature. The first class of methods is built on the theoretical ground that an efficient sample allocation should follow some conditions of sample allocations that asymptotically maximize the PCS. They include a set of equations governing the numbers of samples that should be allocated to the best design and each non-best one. These conditions were first derived in \cite{chen2000a} based on an approximation of PCS using the Bonferroni inequality \cite{bonferroni1936}. A selection procedure was subsequently designed based on the conditions, called the optimal computing budget allocation (OCBA). After that, Glynn and Juneja \cite{glynn2004} analyzed the probability of false selection (PFS, equals 1-PCS) using the large-deviations approach and identified the conditions that optimize the rate function of PFS, which turned out the same as those in \cite{chen2000a} under certain mild assumptions. It suggests that the Bonferroni inequality approximation made to PCS in \cite{chen2000a} is minor in the asymptotic sense and can be ignored. Gao et al. \cite{gao2017} later showed that the same optimality conditions also asymptotically optimize EOC. Following \cite{chen2000a} and \cite{glynn2004}, some variants of R\&S have been considered as well, e.g., subset selection \cite{chen2008,gao2015auto}, multi-objective R\&S \cite{lee2010auto}, complete ranking \cite{xiao2013}, feasibility determination \cite{gao2016feas}, etc.

The second class of methods was among the most important efforts for solving R\&S. They are based on a simple greedy heuristic that iteratively allocates one or a small number of sample(s) to one of the designs in order to optimize the objective measures under study in a greedy manner. Since objective measures such as PCS and EOC do not have analytical expressions in general, early solution approaches simply replaced them with some analytical approximations that could be easily computed. We call them myopic allocation procedures (MAP). Three approximated objective measures commonly used in MAP are the approximated PCS based on the Bonferroni inequality (APCS-B, \cite{Inoue1999,chick2010}), approximated EOC based on the Bonferroni inequality (AEOC-B, \cite{he2007}), and approximated PCS based on the Slepian's inequality (APCS-S, \cite{chen1996,Inoue1999}). Chick et al. \cite{chick2010} summarized these three MAPs and suggested ways to empirically improve their small-sample empirical performance. The idea of myopic allocation was then extended to the value of information procedures (VIP) for selecting the best design. VIPs are not specifically designed for optimizing PCS or EOC. They iteratively allocate samples using predictive distributions of further samples based on the maximizer of a certain acquisition function (e.g., expected improvement and knowledge gradient) that is expected to well balance the exploration of the entire design space and exploitation of the high-quality designs \cite{frazier2008,ryzhov2016}.

MAPs are intuitively appealing and easy to implement, but lack theoretical support compared to other popular R\&S methods. It raises at least two immediate concerns for practitioners to adopt them. First, it is unclear if they are consistent, i.e., the number of samples allocated to each design will go to infinity as the allocation proceeds, so that the estimated best design converges to the true best one and the objective measures such as PCS or EOC converge to 1 or 0. Note that the simple equal allocation (allocating an equal number of samples to each design) is consistent, and will outperform procedures that are not in the long run. Second, it is unclear if the sample allocations generated from MAPs satisfy the optimality conditions established for R\&S \cite{chen2000a,glynn2004}, which serve as an important criterion for judging the asymptotic optimality of sample allocations.

On the other hand, MAPs have been broadly tested using benchmark problems and, surprisingly, demonstrate empirical performance that is competitive to procedures delicately designed based on those optimality conditions in certain scenarios \cite{chen1996,Inoue1999,he2007,chick2010}. In this research, we theoretically study the performance of the three MAPs based on APCS-B, AEOC-B and APCS-S. To the best of our knowledge, it is the first theoretical treatment of R\&S heuristic of myopically optimizing the three objective measure approximations (APCS-B, AEOC-B and APCS-S), after the efforts in \cite{chick2010} for numerically understanding them. We dispel the two concerns mentioned above by showing that the three MAPs, despite their greedy manner, approximations made to the objective measures and the different structures between themselves, all converge to the same sample allocation that satisfies the optimality conditions (note that APCS-B and APCS-S Procedures appeared in the literature before the optimality conditions were developed). In other words, asymptotically speaking, the three MAPs are as good as OCBA, capable of driving the PCS and EOC to converge at the fastest rate. It explains the excellent empirical performance of them. Moreover, it shows the potential of procedures of this type, and indicates a possibly promising future research direction of developing superior MAPs using new approximations of the objective measures, e.g., as in \cite{peng2017}.

There are two streams of related literature that are also established on the goal of selecting the best design from a finite set. The first stream is the indifference-zone method \cite{kim2001,nelson2001}. Instead of optimizing the quality of the selected design, it targets to provide a guarantee on it, and thus leads to different problem structures and methodologies. To build a valid bound for the objective measure, this method often assumes that the mean performance of the best design is at least $\delta$ better than the rest, where $\delta$ is the minimum difference worth detecting. The second stream is the multi-armed bandit (MAB, \cite{Bubeck2012}), which is a reinforcement learning problem originated from \cite{robbins1952}. It mostly considers an online objective measure (i.e., the measurements are associated with the outcome of each sample allocation), while R\&S deals with offline objective measures (i.e., the measurements are associated with only the final estimates). Although recently MAB has been extended to a similar offline setting as R\&S known as the pure exploration or the best arm identification \cite{Audibert2010}, it focuses on the convergence behavior of the procedure, e.g., convergence rate of the simple regret, while R\&S usually concerns with the optimality of the number of samples allocated to each design by the procedure, i.e., the sample allocation, as is in this research.

The remainder of the paper is organized as follows. Section 2 introduces some preliminaries. Section 3 presents the main theoretical results. Section 4 numerically tests the MAPs. Section 5 concludes this paper.

\section{Preliminaries}

In this section, we review the optimality conditions established for R\&S, and describe the APCS-B, AEOC-B and APCS-S Procedures under study in this paper.

\subsection{Optimality Conditions}\label{opt-con}

Suppose we want to select the best from $M$ designs $\left\{1, \ldots, M\right\}$, where best is defined as the design with the largest mean. The mean of each design is unknown and has to be learned by sampling under observation noises. Samples $W_{i}^{(1)}, W_{i}^{(2)}, \dots$ of each design $i$ follow normal distribution with mean $\mu_i$ and variance $\sigma_i^{2}$. Let $N_i$ denote the number of samples allocated to design $i$, and $N=\sum_{i=1}^{M}N_i$ denote the total sampling budget. We assume that the samples are identically and independently distributed from replication to replication for the same design and independent across different designs. The sample mean and sample variance are given by $\hat{\mu}_{i}=\frac{1}{N_i}\sum_{m=1}^{N_i}W_{i}^{\left(m\right)}$, $\hat{\sigma}^{2}_{i}=\frac{1}{N_i-1}\sum_{m=1}^{N_i}\left(W_{i}^{\left(m\right)}-\hat{\mu}_i\right)^{2}$.
Denote by $\hat{b}$ the estimated best design, $\hat{b}=\mathop{\arg\max}_{i}\hat{\mu}_i$. Let $\alpha_i=N_i/N$ be the sample allocation of design $i$, $\sum_{i=1}^{M}\alpha_{i}=1$.

The probability of correct selection (PCS) for the best design is the probability that the estimated best design $\hat{b}$ equals the true best design $b$, i.e., $\text{PCS}=\mathbb{P}\left(\hat{b}=b\right)$. The expected opportunity cost (EOC) is defined as the expectation of the opportunity cost, which is the difference in means between the true best design $b$ and the estimated best design $\hat{b}$, i.e., $\text{EOC}=\mathbb{E}\left[\mu_{b}-\mu_{\hat{b}}\right]$. PCS and EOC are functions of sample allocation $N_i~(i=1,\dots,M)$. To optimize the selection quality, we want to determine the $N_1$, $\dots$, $N_{M}$ that lead to the maximal PCS or the minimal EOC, subject to a fixed sampling budget. In other words, we want to solve the following optimization problems:
\begin{align}
&\max_{N_1,\dots,N_{M}}~\text{PCS}~~
\text{s.t.}~\sum_{i=1}^{M}N_i=N,~N_i\geq 0~\text{for}~\forall i.\label{opt-pcs}\\
&\min_{N_1,\dots,N_{M}}~\text{EOC}~~
\text{s.t.}~\sum_{i=1}^{M}N_i=N,~N_i\geq 0~\text{for}~\forall i.\label{opt-eoc}
\end{align}
In these formulations, we ignore the minor technicalities associated with $N_i$ not being integer. Objective functions PCS and EOC in (\ref{opt-pcs}) and (\ref{opt-eoc}) do not have analytical expressions. To handle this difficulty, a line of research in the literature applied convenient approximations of them in order to solve the optimization programs.

Chen et al. \cite{chen2000a} replaced the objective function PCS in (\ref{opt-pcs}) by an analytical PCS approximation. By investigating the KKT conditions and assuming $N_{b}\gg N_i$ for all $i\ne b$, they derived the optimal computing budget allocation (OCBA) principles: \emph{As $N\to\infty$, the approximated PCS can be maximized if}
\begin{equation}\label{chen-ocba}
\frac{N_{b}^{2}}{\sigma_{b}^{2}}-\sum_{i\ne b}\frac{N_i^{2}}{\sigma_{i}^{2}}=0,~\frac{N_i}{N_j}=\frac{\sigma_i^{2}\left(\mu_j-\mu_{b}\right)^{2}}{\sigma_j^{2}\left(\mu_i-\mu_{b}\right)^{2}},\forall i,j\ne b.
\end{equation}
Glynn and Juneja \cite{glynn2004} studied the optimization problem (\ref{opt-pcs}) using the large-deviations approach \cite{dembo1992}. Denote by $M_i(\gamma)=\log\left(\mathbb{E}[e^{t W_{i}^{\left(m\right)}}]\right)$ the log-moment generating function of $W_{i}^{\left(m\right)}$, by $\Lambda_i(x)=\sup_{\gamma\in\mathbb{R}}\left(\gamma x-M_i(\gamma)\right)$ the Fenchel-Legendre transform of $M_i(\gamma)$, $\forall i$. Note that $\hat{\mu}_i$ depends on the number of samples $N_i=\alpha_i N$ allocated to design $i$. For each vector $V_N=\left(\hat{\mu}_{i}\left(\alpha_i N\right),\hat{\mu}_{b}\left(\alpha_{b} N\right)\right)$, $i\ne b$, we can obtain its rate function $R_i\left(\alpha_i,\alpha_{b}\right)=\inf_{x}\left(\alpha_i\Lambda_i(x)+\alpha_{b}\Lambda_{b}(x)\right)$. Notice that $\lim_{N\to\infty}\frac{1}{N}\log\left(1-\text{PCS}\right)=-\min_{i\ne b}R_i\left(\alpha_i,\alpha_{b}\right)$. Then, the problem (\ref{opt-pcs}) can be re-formulated as 
\begin{equation}\label{opt-pcs-rate}
\begin{aligned}
&\underset{\alpha_1,\dots,\alpha_{M}}{\max}\underset{i\ne b}{\min}~R_i\left(\alpha_i,\alpha_{b}\right)\\
\text{s.t.}&\sum_{i=1}^{M}\alpha_i=1,~\alpha_i\geq 0~\text{for}~\forall i.
\end{aligned}
\end{equation}
Solving it with $W_i^{(m)}\sim\mathcal{N}\left(\mu_i,\sigma_i^{2}\right)$, $\forall i$, $\forall m\in\mathbb{Z}^{+}$, the optimal sample allocation $\alpha_1^{*},\dots,\alpha_M^{*}$ satisfies that
\begin{equation}\label{ocba-nor}
\frac{\left(\alpha_{b}^{*}\right)^{2}}{\sigma_{b}^{2}}-\sum_{i\ne b}\frac{\left(\alpha_i^{*}\right)^{2}}{\sigma_i^{2}}=0,~\frac{\left(\mu_{b}-\mu_i\right)^{2}}{\frac{\sigma_i^{2}}{\alpha_i^{*}}+\frac{\sigma_{b}^{2}}{\alpha_{b}^{*}}}=\frac{\left(\mu_{b}-\mu_j\right)^{2}}{\frac{\sigma_j^{2}}{\alpha_j^{*}}+\frac{\sigma_{b}^{2}}{\alpha_{b}^{*}}},i,j\ne b.
\end{equation}
Gao et al. \cite{gao2017} further showed that asymptotic minimization of EOC in (\ref{opt-eoc}) calls for the same optimality conditions (\ref{ocba-nor}). Note that if we assume $\alpha_{b}^{*}\gg \alpha_i^{*}$ for $\forall i\ne b$ as in \cite{chen2000a}, the equations (\ref{ocba-nor}) become identical to OCBA principles (\ref{chen-ocba}). In this research, we will focus on optimality conditions (\ref{ocba-nor}), which are the solution of the target optimization problems (\ref{opt-pcs}) and (\ref{opt-eoc}) in the asymptotic sense (as $N\rightarrow\infty$) without any additional approximations. We will verify whether the sample allocations generated by the three MAPs satisfy them.

\subsection{Algorithm Description}\label{sec2.2}

In this subsection, we introduce the APCS-B, AEOC-B and APCS-S Procedures. These procedures were first developed under a Bayesian framework. In this research, we follow this setting. For the Bayesian R\&S procedures, unknown parameters are considered as random variables, and their posterior distributions are used to measure the selection quality. Under the assumption of noninformative distributions for the unknown means and variances, the posterior marginal distribution for the unknown mean of design $i$ is a student $t$ distribution with mean $\hat{\mu}_i$, variance $\hat{\sigma}_i^{2}/N_i$ and degree of freedom $N_i-1$, given $N_i>2$ samples \cite{deGroot1970}, $i=1,\dots,M$. Denote by $\tilde{\mu}_i$ the random variable whose distribution is the posterior marginal distribution of unknown $\mu_i$. According to \cite{chick2010}, the difference between $\tilde{\mu}_i$ and $\tilde{\mu}_j$, $i\ne j$, approximately satisfies
\begin{align}\label{diff-means-dist}
\tilde{\mu}_i-\tilde{\mu}_j\sim t\left(\hat{\mu}_i-\hat{\mu}_j,s_{i,j},\nu_{i,j}\right),
\end{align}
where $\nu_{i,j}=s_{i,j}^{2}\big{/}\left(\frac{\hat{\sigma}_{i}^{4}}{N_{i}^{2}\left(N_{i}-1\right)}+\frac{\hat{\sigma}_{j}^{4}}{N_{j}^{2}\left(N_{j}-1\right)}\right)$, $s_{i,j}=\frac{\hat{\sigma}_{i}^{2}}{N_{i}}+\frac{\hat{\sigma}_{j}^{2}}{N_{j}}$. We adopt the noninformative priori of the unknown means and variances throughout the paper.

Given all samples $\mathcal{W}=\left\{\left.W_i^{(1)},\dots,W_i^{\left(N_i\right)}\right|i=1,\dots,M\right\}$ obtained so far, we can transfer PCS and EOC mentioned in Section \ref{opt-con} to a Bayesian form:
\begin{align}
&\text{PCS}_{\text{Bayes}}=\mathbb{P}\left(\left.\max_{i\ne\hat{b}}\tilde{\mu}_i<\tilde{\mu}_{\hat{b}}\right|\mathcal{W}\right),\label{pcs}\\
&\text{EOC}_{\text{Bayes}}=\mathbb{E}\left[\left.\tilde{\mu}_{b}-\tilde{\mu}_{\hat{b}}\right|\mathcal{W}\right].\label{eoc}
\end{align}
Furthermore, some estimation techniques will be used to approximate $\text{PCS}_{\text{Bayes}}$ and $\text{EOC}_{\text{Bayes}}$. The Slepian's inequality \cite{chen1996,Inoue1999} states that $\mathbb{P}\left\{\bigcap_{i=1}^{M}\left(X_i\leq c_i\right)\right\}\geq \prod_{i=1}^{M}\mathbb{P}\left(X_i\leq c_i\right)$, where $\left(X_1,\dots,X_M\right)$ follows an $M$-variate normal distribution with zero mean vector, unit variances and nonnegative correlation coefficients. $c_1$, $\dots$, $c_M$ are some constants.
Applying the Slepian's inequality and (\ref{diff-means-dist}) to (\ref{pcs}), $\text{PCS}_{\text{Bayes}}\geq\prod_{i\ne \hat{b}}\mathbb{P}\left(\left.\tilde{\mu}_i<\tilde{\mu}_{\hat{b}}\right|\mathcal{W}\right)\triangleq \text{APCS-S}$,
\begin{align}
\text{APCS-S}
\doteq\prod_{i\ne \hat{b}}\Phi_{\nu_{i,\hat{b}}}\left(d_{i,\hat{b}}\right),\label{APCS-S}
\end{align}
and we refer this lower bound of $\text{PCS}_{\text{Bayes}}$ as the approximated $\text{PCS}$ using the Slepian's inequality  ($\text{APCS-S}$), where $d_{i,\hat{b}}=\left(\hat{\mu}_{\hat{b}}-\hat{\mu}_{j}\right)\big{/}\sqrt{s_{i,\hat{b}}}$, $s_{i,\hat{b}}$ and $\nu_{i,\hat{b}}$ are from (\ref{diff-means-dist}), $\Phi_{\nu}(x)=\int_{-\infty}^{x}\phi_{\nu}(r)\text{d}r$, $\phi_{\nu}(x)=\frac{1}{\sqrt{\nu}\cdot \text{B}\left(\frac{1}{2},\frac{\nu}{2}\right)}\left(1+\frac{r^{2}}{\nu}\right)^{-\frac{\nu+1}{2}}$, $\text{B}\left(\cdot,\cdot\right)$ denotes the beta function. For more detailed explanation, see \cite{chick2010}. 

The Bonferroni inequality \cite{bonferroni1936,chen2000a} indicates that for events $F_1$, $\dots$, $F_M$, $\mathbb{P}\left(\bigcup_{i=1}^{M}F_i\right)
\leq \sum_{i=1}^{M}\mathbb{P}\left(F_i\right)$. Applying the Bonferroni inequality and (\ref{diff-means-dist}) to (\ref{pcs}), $\text{PCS}_{\text{Bayes}}\geq 1-\sum_{i\ne \hat{b}}\mathbb{P}\left(\left.\tilde{\mu}_i>\tilde{\mu}_{\hat{b}}\right|\mathcal{W}\right)\triangleq \text{APCS-B}$,
\begin{align}
\text{APCS-B}\doteq 1-\sum_{i\ne \hat{b}}\Phi_{\nu_{i,\hat{b}}}\left(-d_{i,\hat{b}}\right),\label{APCS-B}
\end{align}
and we refer this lower bound of $\text{PCS}_{\text{Bayes}}$ as the approximated $\text{PCS}$ using the Bonferroni inequality ($\text{APCS-B}$),
where $\nu_{i,\hat{b}}$,
$d_{i,\hat{b}}$ are defined in (\ref{diff-means-dist}) and (\ref{APCS-S}). 

In addition, we can apply the Bonferroni inequality and (\ref{diff-means-dist}) to $\text{EOC}_{\text{Bayes}}$ in (\ref{eoc}). $\text{EOC}_{\text{Bayes}}\leq\sum_{i\ne \hat{b}}\int_{0}^{+\infty}y\cdot h_{i,\hat{b}}(y)\text{d}y\triangleq \text{AEOC-B}$,
\begin{align}\label{AEOC-B}
\text{AEOC-B}\doteq\sum_{i\ne\hat{b}}s_{i,\hat{b}}^{\frac{1}{2}}\Psi_{\nu_{i,\hat{b}}}\left(d_{i,\hat{b}}\right).
\end{align}
We refer this upper bound of $\text{EOC}_{\text{Bayes}}$ as the approximated $\text{EOC}$ using the Bonferroni inequality ($\text{AEOC-B}$),
where $h_{i,\hat{b}}(y)$ denotes the probability density function of $Y\sim t\left(\hat{\mu}_{\hat{b}}-\hat{\mu}_i,s_{i,\hat{b}},\nu_{i,\hat{b}}\right)$, $s_{i,\hat{b}}$, $\nu_{i,\hat{b}}$ and $d_{i,\hat{b}}$ are defined in (\ref{diff-means-dist}) and (\ref{APCS-S}), and $\Psi_{\nu}(x)=\frac{\nu+x^{2}}{\nu-1}\phi_{\nu}(x)-x\Phi_{\nu}(-x)$.

Given the budget of $N$ samples, let $\left\{I^{(n)}\right\}_{n=1}^{N}$ be the sequence of designs chosen by the procedure for sampling. The quantity $n$ is the iteration index of the procedure. Denote by $N_i^{\left(n\right)}$ the number of samples assigned to $i$ up to iteration $n$. $N_i^{\left(n\right)}=\sum_{m=1}^{n}\mathbbm{1}{\{I^{(m)}=i\}}$, where $\mathbbm{1}\left\{\cdot\right\}$ is an indicator function. Sample mean, sample variance and the estimated best design in iteration $n$ are $\hat{\mu}^{\left(n\right)}_{i}=\frac{1}{N_i^{\left(n\right)}}\sum_{m=1}^{N_i^{\left(n\right)}}W_{i}^{\left(m\right)}$,	 $\left(\hat{\sigma}^{\left(n\right)}_{i}\right)^{2}=\frac{1}{N_i^{\left(n\right)}-1}\sum_{m=1}^{N_i^{\left(n\right)}} \left(W_{i}^{\left(m\right)}-\hat{\mu}_i^{\left(n\right)}\right)^{2}$, $\hat{b}^{(n)}=\arg\max_i\hat{\mu}_{i}^{(n)}$. The three approximation measures can be calculated by
\begin{align}\label{apcsb}
\text{APCS-B}^{\left(n\right)}=1-\sum_{i\ne \hat{b}^{(n)}}\Phi_{\nu_{i,\hat{b}^{(n)}}^{\left(n\right)}}\left(-d_{i,\hat{b}^{(n)}}^{\left(n\right)}\right),
\end{align}
\begin{align}
\text{AEOC-B}^{\left(n\right)}
&=\sum_{i\ne\hat{b}^{(n)}}\sqrt{s_{i,\hat{b}^{(n)}}^{\left(n\right)}}\cdot\Psi_{\nu_{i,\hat{b}^{(n)}}^{\left(n\right)}}\left(d_{i,\hat{b}^{(n)}}^{\left(n\right)}\right),\label{aeocb}\\
\text{APCS-S}^{\left(n\right)}
&=\prod_{i\ne \hat{b}^{(n)}}\Phi_{\nu_{i,\hat{b}^{(n)}}^{\left(n\right)}}\left(d_{i,\hat{b}^{(n)}}^{\left(n\right)}\right).\label{apcss}
\end{align}
To evaluate the effect of more sampling on $\text{APCS-B}^{\left(n\right)}$, $\text{AEOC-B}^{\left(n\right)}$ and $\text{APCS-S}^{\left(n\right)}$, suppose one more sample is allocated to design $j$ with $N_j^{(n)}$ increased to $N_j^{(n)}+1$. We re-calculate the sample mean $\hat{\mu}_j^{(n)}$, sample variance $\left(\hat{\sigma}_j^{(n)}\right)^{2}$ and the estimated best design $\hat{b}^{(n)}$. For $i\ne \hat{b}^{(n)}$, parameters $s_{i,\hat{b}^{(n)}}^{\left(n\right)}$, $d_{i,\hat{b}^{(n)}}^{\left(n\right)}$ and $\nu_{i,\hat{b}^{(n)}}^{\left(n\right)}$ are updated as
\begin{flalign}
\tilde{s}_{i,\hat{b}^{(n)}}^{\left(n\right),j}=&
\frac{\left(\hat{\sigma}_{i}^{\left(n\right)}\right)^{2}}{N^{\left(n\right)}_{i}+\mathbbm{1}\{j=i\}}+\frac{\left(\hat{\sigma}_{\hat{b}^{(n)}}^{\left(n\right)}\right)^{2}}{N^{\left(n\right)}_{\hat{b}^{(n)}}+\mathbbm{1}\left\{j=\hat{b}^{(n)}\right\}},\nonumber\\
\tilde{d}^{\left(n\right),j}_{i,\hat{b}^{(n)}}=&\left(\hat{\mu}^{\left(n\right)}_{\hat{b}^{(n)}}-\hat{\mu}^{\left(n\right)}_{i}\right)\Big{/}\sqrt{\tilde{s}_{i,\hat{b}^{(n)}}^{\left(n\right),j}},\label{tilde_d&s&nu_n}\\
\tilde{\nu}^{\left(n\right),j}_{i,\hat{b}^{(n)}}=&\frac{\left(\tilde{s}_{i,\hat{b}^{(n)}}^{\left(n\right),j}\right)^{2}}
{\frac{\left(\frac{\left(\hat{\sigma}_{i}^{\left(n\right)}\right)^{2}}{N^{\left(n\right)}_{i}
			+\mathbbm{1}\{j=i\}}\right)^{2}}{N^{\left(n\right)}_{i}-1+\mathbbm{1}\{j=i\}}
	+\frac{\left(\frac{\left(\hat{\sigma}_{\hat{b}^{(n)}}^{\left(n\right)}\right)^{2}}{N^{\left(n\right)}_{\hat{b}^{(n)}}+\mathbbm{1}\left\{j=\hat{b}^{(n)}\right\}}\right)^{2}}{N^{\left(n\right)}_{\hat{b}^{(n)}}-1+\mathbbm{1}\left\{j=\hat{b}^{(n)}\right\}}}.\nonumber
\end{flalign}
Let $\text{APCS-B}^{\left(n\right),j}=1-\sum_{i\ne \hat{b}^{(n)}}\Phi_{\tilde{\nu}_{i,\hat{b}^{(n)}}^{\left(n\right),j}}\left(-\tilde{d}_{i,\hat{b}^{(n)}}^{\left(n\right),j}\right)$ be the value of $\text{APCS-B}^{(n)}$ after design $j$ receives one more sample, 
and $I^{\left(n\right)}=\mathop{\arg\max}_{i} \left[\text{APCS-B}^{\left(n\right),i}-\text{APCS-B}^{\left(n\right)}\right]$ be the design that improves $\text{APCS-B}^{(n)}$ the most after sampling. The APCS-B Procedure is shown below.
\begin{framed}[0.8\textwidth]
	\textbf{APCS-B Procedure}
	\begin{itemize}
		\item[1:] Initialize the number of designs $M$ and the sampling budget $N$.
		\item[2:] Collect $N_0(\ll N/M)$ samples for each $i$.
		\item[3:] $n\leftarrow 0$, $N_1^{(n)}=\cdots=N_M^{(n)}=N_0$.
		\item[4:] \textbf{WHILE} $\sum_{i=1}^{M}N_i^{(n)}<N$ \textbf{DO}
		\item[5:] Update $\hat{\mu}_i^{(n)}$ and $\hat{\sigma}_i^{(n)}$ for each $i$, and $\hat{b}^{(n)}$.
		\item[6:] $I^{\left(n\right)}=\mathop{\arg\max}_i \left[\text{APCS-B}^{\left(n\right),i}-\text{APCS-B}^{\left(n\right)}\right]$.
		\item[7:] Collect one more sample for design $I^{\left(n\right)}$.
		\item[8:] Update $N_{i}^{(n+1)}=N_{i}^{(n)}+\mathbbm{1}\left\{I^{\left(n\right)}=i\right\}$ for each $i$. $n\leftarrow n+1$.
		\item[9:] \textbf{END WHILE}
		\item[10:] Select $\hat{b}^{\left(N\right)}$ as the estimated best design.
	\end{itemize}
\end{framed}

In the initialization stage (lines 1-3 of the APCS-B Procedure), $N_0(\ll N/M)$ samples are allocated to each of the $M$ designs to obtain initial estimates for the means, variances and the best design. In each of the subsequent iterations, the procedure identifies the design $I^{\left(n\right)}$ that improves $\text{APCS-B}^{(n)}$ the most after receiving one more sample, allocates one sample to $I^{\left(n\right)}$ and updates relevant estimators. This procedure is continued until the sampling budget $N$ is exhausted. In the end, the procedure outputs the design with the largest sample mean.

\begin{remark}
	Although both the OCBA \cite{chen2000a} and APCS-B \cite{Inoue1999,chick2010} Procedures are based on the Bonferroni inequality, they are significantly different. The OCBA Procedure was built based on the optimality conditions, which were obtained by using the global optimization method to maximize an approximation of PCS (APCS). In contrast, the APCS-B Procedure applies the myopic optimization method to maximize the APCS. That is, the procedure iteratively allocates samples to the designs which improve the APCS the most. It is widely known that the myopic optimization algorithm might not always find the global optimum \cite{black2005}. In this research, we theoretically prove that the sample allocations generated by the myopic R\&S procedures (e.g., the APCS-B Procedure) can converge to the optimal allocation calculated by the optimality conditions, which is the global optimum of the problem (\ref{opt-pcs-rate}). \hfill $\square$
\end{remark}

For the AEOC-B measure, we can assess the effect of one more sample for design $j$. Let $\text{AEOC-B}^{\left(n\right),j}$ be the value of $\text{AEOC-B}^{(n)}$ after design $j$ receives one more sample.
\begin{equation}\label{aeocbj}
\text{AEOC-B}^{\left(n\right),j}=
\sum_{i\ne \hat{b}^{(n)}}
\sqrt{\tilde{s}_{i,\hat{b}^{(n)}}^{\left(n\right),j}}\cdot\Psi_{\tilde{\nu}^{\left(n\right),j}_{i,\hat{b}^{(n)}}}\left(\tilde{d}^{\left(n\right),j}_{i,\hat{b}^{(n)}}\right),
\end{equation}
where terms $\tilde{s}_{i,\hat{b}^{(n)}}^{\left(n\right),j}$, $\tilde{d}_{i,\hat{b}^{(n)}}^{\left(n\right),j}$ and $\tilde{\nu}_{i,\hat{b}^{(n)}}^{\left(n\right),j}$ are the same as in (\ref{tilde_d&s&nu_n}). The AEOC-B Procedure is similar to the APCS-B Procedure in the myopic manner. It iteratively samples the design that reduces $\text{AEOC-B}^{\left(n\right)}$ the most. The AEOC-B Procedure is summarized below.
\begin{framed}[0.75\textwidth]
	\textbf{AEOC-B Procedure}
	\begin{itemize}
		\item[1:] Initialize $M$ and $N$.
		\item[2:] Collect $N_0(\ll N/M)$ samples for each $i$.
		\item[3:] $n\leftarrow 0$, $N_1^{(n)}=\cdots=N_M^{(n)}=N_0$.
		\item[4:] \textbf{WHILE} $\sum_{i=1}^{M}N_i^{(n)}<N$ \textbf{DO}
		\item[5:] Update $\hat{\mu}_i^{(n)}$ and $\hat{\sigma}_i^{(n)}$ for each $i$, and $\hat{b}^{(n)}$.
		\item[6:] $I^{\left(n\right)}=\mathop{\arg\max}_i \left[\text{AEOC-B}^{\left(n\right)}-\text{AEOC-B}^{\left(n\right),i}\right]$.
		\item[7:] Collect one more sample for design $I^{\left(n\right)}$.
		\item[8:] Update $N_{i}^{(n+1)}=N_{i}^{(n)}+\mathbbm{1}\left\{I^{\left(n\right)}=i\right\}$ for each $i$. $n\leftarrow n+1$.
		\item[9:] \textbf{END WHILE}
		\item[10:] Select $\hat{b}^{\left(N\right)}$ as the estimated best design.
	\end{itemize}
\end{framed}

For the APCS-S measure, we can also assess the effect of one more sample for design $j$. Let $\text{APCS-S}^{\left(n\right),j}$ be the value of $\text{APCS-S}^{(n)}$ after design $j$ receives one more sample.
\begin{equation}\label{apcssj}
\text{APCS-S}^{\left(n\right),j}=
\prod_{i\ne \hat{b}^{(n)}}
\Phi_{\tilde{\nu}^{\left(n\right),j}_{i,\hat{b}^{(n)}}}\left(\tilde{d}^{\left(n\right),j}_{i,\hat{b}^{(n)}}\right),
\end{equation}
where terms $\tilde{s}_{i,\hat{b}^{(n)}}^{\left(n\right),j}$, $\tilde{d}_{i,\hat{b}^{(n)}}^{\left(n\right),j}$ and $\tilde{\nu}_{i,\hat{b}^{(n)}}^{\left(n\right),j}$ are the same as in (\ref{tilde_d&s&nu_n}). The APCS-S Procedure iteratively samples the design that improves $\text{APCS-S}^{\left(n\right)}$ the most. It is summarized below.

\begin{framed}[0.75\textwidth]
	\textbf{APCS-S Procedure}
	\vspace*{-5pt}
	\begin{itemize}
		\item[1:] Initialize $M$ and $N$.
		\item[2:] Collect $N_0(\ll N/M)$ samples for each $i$.
		\item[3:] $n\leftarrow 0$, $N_1^{(n)}=\cdots=N_M^{(n)}=N_0$.
		\item[4:] \textbf{WHILE} $\sum_{i=1}^{M}N_i^{(n)}<N$ \textbf{DO}
		\item[5:] Update $\hat{\mu}_i^{(n)}$ and $\hat{\sigma}_i^{(n)}$ for each $i$, and $\hat{b}^{(n)}$.
		\item[6:] $I^{\left(n\right)}=\mathop{\arg\max}_i \left[\text{APCS-S}^{\left(n\right),i}-\text{APCS-S}^{\left(n\right)}\right]$.
		\item[7:] Collect one more sample for design $I^{\left(n\right)}$.
		\item[8:] Update $N_{i}^{(n+1)}=N_{i}^{(n)}+\mathbbm{1}\left\{I^{\left(n\right)}=i\right\}$ for each $i$. $n\leftarrow n+1$.
		\item[9:] \textbf{END WHILE}
		\item[10:] Select $\hat{b}^{\left(N\right)}$ as the estimated best design.
	\end{itemize}
\end{framed}

\begin{remark}
	The three MAPs work on the same naive idea. To maximize the PCS in (\ref{opt-pcs}) or the EOC in (\ref{opt-eoc}) under a sampling budget of $N$, they simply use one sample at a time to myopically optimize an approximation of PCS or EOC until the $N$ samples are used up. This selection mechanism of the three procedures can be attractive to decision-makers because they possess simple rationales and concise structures and are easy to implement. Note that even for an optimization problem with no complication from approximation and sample randomness, the greedy improvement heuristic does not guarantee optimality in general. However, as can be seen later, this heuristic turns out asymptotically optimal for the optimization problems (\ref{opt-pcs}) and (\ref{opt-eoc}), and is thus an appealing method for solving R\&S. \hfill $\square$
\end{remark}

\section{Analysis of the Myopic Allocation Procedures}\label{sec3}

In this section, we characterize the theoretical performance of the APCS-B, AEOC-B and APCS-S Procedures.

Lemma \ref{lem3} below shows a convergence property of real sequences. It will be used in the proof of Theorem \ref{thm-optsampalloc}. The detailed proof of Lemma \ref{lem3} is shown in Appendix \ref{app_lem1}.
\begin{lemma}\label{lem3}
	Let $\left\{N_i^{(n)}\Big{|}i=1,\dots,M,n=1,2,\dots\right\}$ be a sequence of positive integers that satisfy $N_i^{(n)}\to\infty$ as $n\to\infty$, where $M\geq 2$ is also a positive integer. Let $\alpha_i^{(n)}=N_i^{(n)}/\sum_{j=1}^{M}N_j^{(n)}$ and $L_i^{(n)}=L_{i}^{(n-1)}+\mathbbm{1}\left\{N_i^{(n)}>N_i^{(n-1)}\right\}$ for $\forall i$, where $\mathbbm{1}\left\{\cdot\right\}$ is an indicator function and $L_i^{(0)}=0$ for $\forall i$. For the sequence $\left\{\alpha_i^{(n)}\Big{|}i=1,\dots,M,n=1,2,\dots\right\}$, if each subsequence $\left\{\rule{0em}{4mm}\alpha_i^{\left(n_t\right)}\Big{|}t=1,2,\dots,L_i^{\left(t\right)}\to\infty\right.$\\$\left.~\text{as}~t\to\infty,i=1,\dots,M\rule{0em}{4mm}\right\}$ has a convergent subsequence $\left\{\rule{0em}{4.5mm}\alpha_i^{\left(n_{t_q}\right)}\Big{|}q=1,2,\dots,L_i^{\left(q\right)}\to\infty\text{as}~q\to\infty,\right.$\\$\left.i=1,\dots,M\rule{0em}{4.5mm}\right\}$ and the limit of the convergent subsequence satisfies the optimality conditions (\ref{ocba-nor}), then  $\left\{\alpha_i^{(n)}\Big{|}i=1,\dots,M,n=1,2,\dots\right\}$ satisfies
	\begin{equation}\label{lem3-eq1}
	\begin{aligned}
	\lim_{n\to\infty}\left(\frac{\alpha_b^{\left(n\right)}}{\sigma_b}\right)^{2}-\sum_{i\ne b}\left(\frac{\alpha_i^{\left(n\right)}}{\sigma_i}\right)^{2}=0,~
	\lim_{n\to\infty}\frac{\left(\mu_b-\mu_i\right)^{2}}{\frac{\sigma_i^{2}}{\alpha_i^{\left(n\right)}}+\frac{\sigma_b^{2}}{\alpha_b^{\left(n\right)}}}-\frac{\left(\mu_b-\mu_j\right)^{2}}{\frac{\sigma_j^{2}}{\alpha_j^{\left(n\right)}}+\frac{\sigma_b^{2}}{\alpha_b^{\left(n\right)}}}=0,~\forall i,j\ne b.
	\end{aligned}
	\end{equation}
\end{lemma}

To obtain our main results, we first consider the consistency of the three MAPs. Under a Bayesian framework, consistency is a property that the posterior belief converges to the underlying truth. In the R\&S problem, means of the designs and the best design $b$ are unknown. With consistency of the procedures, the posterior distributions of $\tilde{\mu}_i$ tend to degenerate distributions concentrated on a single point as the total number of samples goes to infinity. In this case, the MAPs almost surely output the best design $b$.

\begin{theorem}\label{thm-consistency}
	For the APCS-B, AEOC-B and APCS-S Procedures, $\lim_{n\to\infty}\hat{b}^{\left(n\right)}\overset{a.s.}{=}b$.
\end{theorem}

In addition to the APCS-B, AEOC-B and APCS-S Procedures, it is well known that a lot of other R\&S algorithms are consistent. For example, the simple equal allocation allocates an equal number of samples to each design; for the OCBA algorithm \cite{chen2000a}, the number of samples allocated to each design is proportional to the sampling budget. These sample allocations have clear long-term exploration mechanisms that ensure sufficient exploration on bad designs. With them, the consistency property immediately follows. In contrast, purely myopic algorithms only seek short-term optimization of the objective functions, so consistency of myopic algorithms cannot be guaranteed in general. Theorem \ref{thm-consistency} shows the consistency of the three MAPs under study in this paper. It suggests that the approximations APCS-B, AEOC-B and APCS-S possess good structures for guiding sample allocations, which provide sufficient number of samples to bad designs.

The detailed proof of Theorem \ref{thm-consistency} is shown in Appendix \ref{app_thm1}. Below we give a proof sketch for Theorem \ref{thm-consistency}. Take the APCS-B Procedure as an example. To show its consistency, it suffices to prove that the number of samples allocated to each design goes to infinity as the budget goes to infinity. We show it by contradiction. We can derive that for design $i$ whose number of samples is finite in the limit of $n\to\infty$, $\lim_{n\to\infty}\left[\text{APCS-B}^{(n),i}-\text{APCS-B}^{(n)}\right]>0$; meanwhile, for design $j\ne i$ whose number of samples goes to infinity in the limit of $n\to\infty$, $\lim_{n\to\infty}\left[\text{APCS-B}^{(n),j}-\text{APCS-B}^{(n)}\right]=0$. It indicates that $\exists\zeta$, $\forall n>\zeta$, $\text{APCS-B}^{(n),i}-\text{APCS-B}^{(n)}>\text{APCS-B}^{(n),j}-\text{APCS-B}^{(n)}$. Notice that the APCS-B Procedure samples the design that improves $\text{APCS-B}^{(n)}$ the most in iteration $n$, i.e., $I^{\left(n\right)}=\mathop{\arg\max}_{i} \left[\text{APCS-B}^{\left(n\right),i}-\text{APCS-B}^{\left(n\right)}\right]$. In this case, the APCS-B Procedure would prefer sampling design $i$ to design $j$ in iterations $n>\zeta$. It means that $j$ will no longer be sampled after $n>\zeta$ and will receive at most a finite number of samples as $n\to\infty$, which contradicts that the number of samples of design $j$ goes to infinity. Thus, the number of samples of each design goes to infinity as budget goes to infinity. The detailed proof of Theorem \ref{thm-consistency} is provided as follows.

%\begin{remark}
%	In addition to the APCS-B, AEOC-B and APCS-S Procedures, it is well known that a lot of other R\&S algorithms are consistent. For example, the simple equal allocation allocates an equal number of samples to each design; for the OCBA Algorithm \cite{chen2000a}, the number of samples allocated to each design is proportional to the sampling budget. These sample allocations have clear long-term exploration mechanisms that ensure sufficient exploration on bad designs. With them, the consistency property immediately follows. In contrast, purely myopic algorithms only seek short-term optimization of the objective functions, so consistency of myopic algorithms cannot be guaranteed in general. Theorem \ref{thm-consistency} shows the consistency of the three MAPs under study in this paper. It suggests that the approximations APCS-B, AEOC-B and APCS-S possess good structures for guiding sample allocations, which provide sufficient number of samples to bad designs.
%	\hfill $\square$
%\end{remark}

In addition to consistency, another major concern for the MAPs is whether their sample allocations satisfy the optimality conditions (\ref{ocba-nor}) or not. If the sample allocations from the MAPs satisfy (\ref{ocba-nor}), the sample allocations generated from the three MAPs solve (\ref{opt-pcs}) and (\ref{opt-eoc}) in an asymptotic sense.
Inspired by \cite{ryzhov2016} and \cite{peng2018tac}, we show that the sample allocations from the three MAPs satisfy (\ref{ocba-nor}) as the budget goes to infinity.

\begin{theorem}\label{thm-optsampalloc}
	The sample allocations of the APCS-B, AEOC-B and APCS-S Procedures converge to the solution of the optimality conditions (\ref{ocba-nor}) almost surely, i.e.,
	\begin{align}
	&\lim_{n\to\infty}\frac{\left(\alpha_{b}^{(n)}\right)^{2}}{\sigma_{b}^{2}}-\sum_{i\ne b}\frac{\left(\alpha_i^{(n)}\right)^{2}}{\sigma_i^{2}}\overset{a.s.}{=}0,\label{thm-opteqs1}\\
	\lim_{n\to\infty}&\frac{\left(\mu_{b}-\mu_i\right)^{2}}{\frac{\sigma_i^{2}}{\alpha_i^{(n)}}+\frac{\sigma_{b}^{2}}{\alpha_{b}^{(n)}}}-\frac{\left(\mu_{b}-\mu_j\right)^{2}}{\frac{\sigma_j^{2}}{\alpha_j^{(n)}}+\frac{\sigma_{b}^{2}}{\alpha_{b}^{(n)}}}\overset{a.s.}{=}0,~\forall i,j\ne b.\label{thm-opteqs2}
	\end{align}
\end{theorem}

The detailed proof of Theorem \ref{thm-optsampalloc} is shown in Appendix \ref{app_thm2}. Below we provide a proof sketch for Theorem 2. Take the APCS-B Procedure as an example. At first, we present a modified version of the APCS-B Procedure in which the sample mean and sample variance of each design are replaced by the true mean and true variance. Then, by similar discussion as used in the proof of Theorem 1, we show that this modified APCS-B Procedure leads the number of samples allocated to each design to go to infinity. Based on that, we prove that the modified APCS-B Procedure asymptotically satisfies optimality conditions (\ref{ocba-nor}). Next, we prove the statement of Theorem 2 for the APCS-B Procedure by showing that the deviation in sample allocations between the APCS-B Procedure and the modified APCS-B Procedure is negligible in the long run, which indicates that the sample allocation from the APCS-B Procedure also satisfies (\ref{ocba-nor}).

\section{Numerical Experiments}\label{sec4}

In this section, we numerically test the three MAPs. For comparison, we also test Equal Allocation (EA) and the OCBA Procedure \cite{chen2000a,li2022b}. Notice that before running the MAP and the OCBA procedures, we need to set the initial number of samples $N_0$ and the one-time computing budget increment $\Delta$. These parameters may influence the performance of the procedures. At first, we consider three numerical examples with different structures. The procedures share the same values of $N_0=2$ and $\Delta=1$. We evaluate the performance of the compared procedures by three measures: sample allocations on some selected designs, PCS and EOC. Then, we discuss the sensitivity of the MAP and OCBA procedures to $N_0$ and  $\Delta$, by comparing the sampling budget needed for the procedures to achieve the desired PCS.

\subsection{Numerical Tests on Different Examples}\label{sec4.1}

In this section, we test the three MAPs and compare them with the OCBA and EA procedures by performing a series of numerical experiments.

\begin{itemize}
	\item Increasing Mean Configuration:\\
	In this experiment, we consider $10$ designs $\left\{1,\dots,10\right\}$. For $i=1,\dots,10$, the observations of design $i$ follow a normal distribution with mean $\mu_i=i$ and standard deviation $\sigma_i$ randomly generated from uniform distribution $U\left(0,6\right)$, $i=1,\dots,10$. The goal is to select the best design $b=10$ with the largest mean.
	
	The numerical results are shown in Figure \ref{fig1}. In Figures \ref{fig1}(i) and \ref{fig1}(ii), the red, green, pale blue, dark blue, and black lines represent the APCS-B Procedure, AEOC-B Procedure, APCS-S Procedure, OCBA Procedure and EA respectively. In Figure \ref{fig1}(iii), black lines represent the optimal sample allocation calculated by (\ref{ocba-nor}). For simplicity, we select three designs from the solution space and present their sample allocation trends. To reduce the influence of randomness in the procedures, each compared procedure is repeated for 500 replications to obtain estimates of PCS and EOC, and is repeated for 20 replications to obtain estimates of sample allocations on the selected designs. These settings remain the same for the three numerical examples unless otherwise specified.
	
	\begin{figure}[!htbp]
		\centering
		\begin{subfigure}
			\centering
			\includegraphics[scale=0.52]{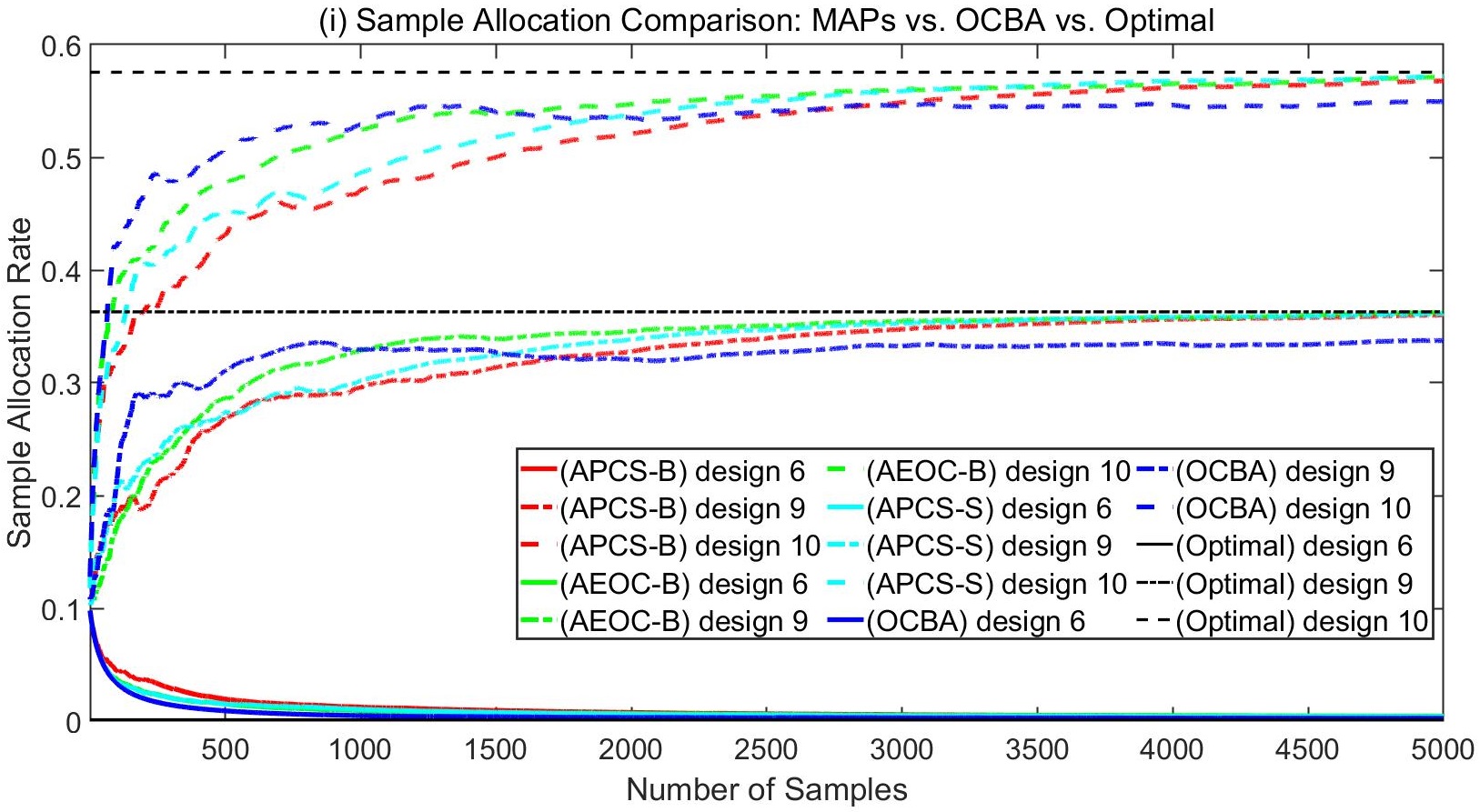}
		\end{subfigure}
		\begin{subfigure}
			\centering
			\includegraphics[scale=0.52]{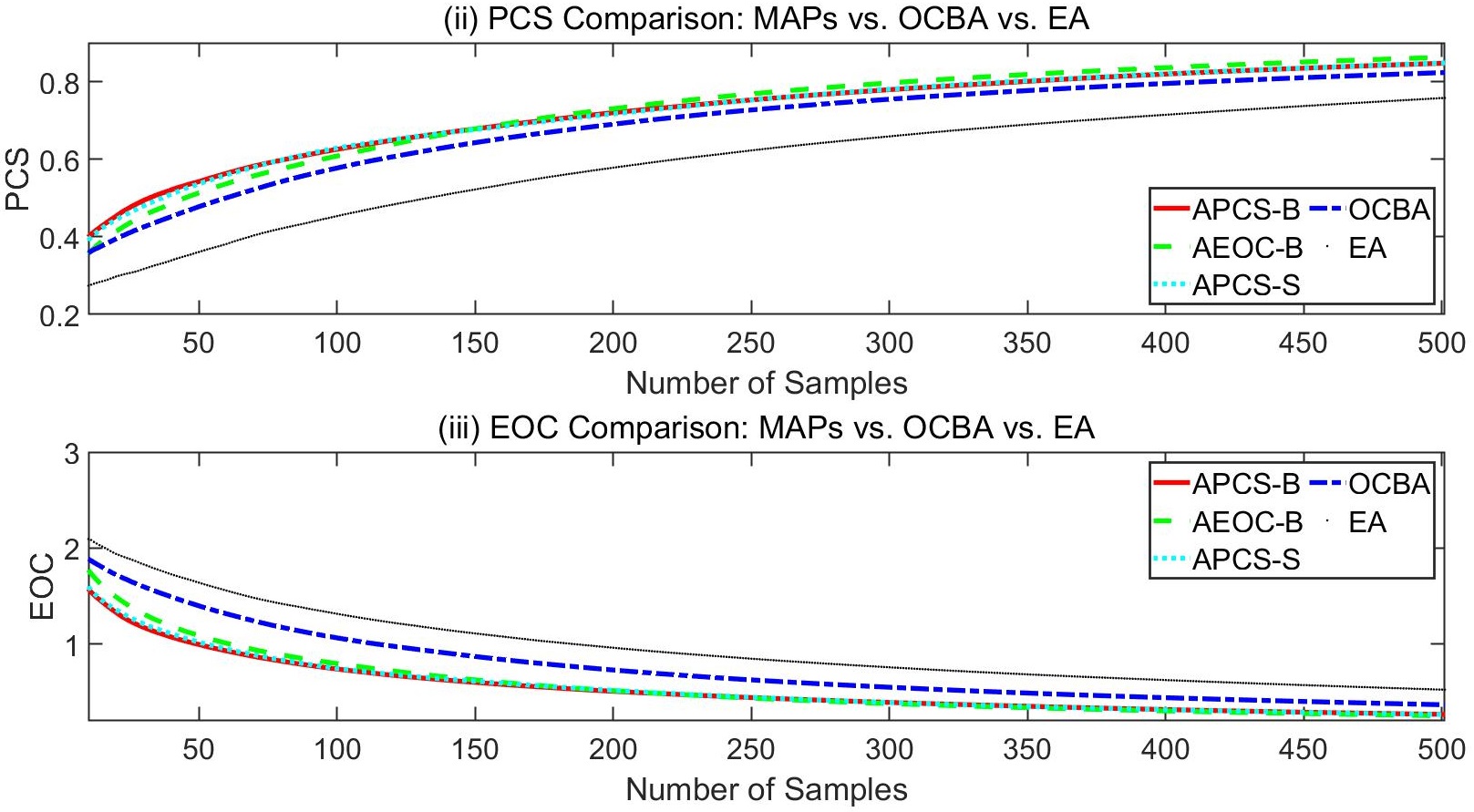}
		\end{subfigure}
		\caption{Increasing Mean Configuration.}
		\label{fig1}
	\end{figure}
	
	%	\begin{figure}[!htbp]
	%		\begin{center}
	%			\includegraphics[scale=0.24]{Figures/figMIM.jpg}
	%			\caption{}
	%			\label{fig1}
	%		\end{center}
	%	\end{figure}
	
	In Figure \ref{fig1}(i), it can be observed that sample allocations of designs 6, 9 and 10 generated by the three MAPs gradually approach their theoretically optimal sample allocations. The OCBA Procedure leads the sample allocations of the designs to values that are slightly different from the theoretically optimal ones in the long run, due to the slight difference between equations (\ref{chen-ocba}) and equations (\ref{ocba-nor}), but when the total number of samples is relatively small, OCBA has an edge over the MAPs. The three MAPs have similar performance as the total number of samples increases. In Figures \ref{fig1}(ii) and \ref{fig1}(iii), it is observed that the MAPs dominate the OCBA Procedure and EA with respect to both PCS and EOC as the procedures proceed. OCBA is generally better than EA, especially in the early stages of the sample allocation. The three MAPs have similar performance under PCS and EOC. 
	
	\item Test Functions:\\
	We next conduct numerical experiments on two common benchmark functions \cite{jamil2013} over discrete solution sets, shown as follows:
	
	\begin{itemize}
		\item Rosenbrock function: \begin{align*}
		    f_1\left(x_1,x_2\right)=\left(x_1-1\right)^{2}+100\left(x_2-x_1^{2}\right)^{2},
		\end{align*}
		where $x_1,x_2\in\left\{-2,-1,0,1,2\right\}$. The unique global minimum is at $\left(x_1^{*},x_2^{*}\right)=\left(1,1\right)$.	
		
		\item Goldstein-Price function:
		\begin{align*}
		    f_2\left(x_1,x_2\right)=\frac{1}{100}&\Big{[}1+\left(x_1+x_2+1\right)^{2}\big{(}19-14x_1+3x_1^{2}-14x_2+6x_1 x_2+3x_2^{2}\big{)}\Big{]}\\
		    &\cdot\Big{[}30+\left(2x_1-3x_2\right)^{2}\big{(}18-32x_1+12x_1^{2}+48x_2-36x_1 x_2+27x_2^{2}\big{)}\Big{]}
		\end{align*}
		where $x_1,x_2\in\left\{-2,-1,0,1,2\right\}$. The unique global minimum is at $\left(x_1^{*},x_2^{*}\right)=\left(0,-1\right)$.		
	\end{itemize}
\end{itemize}

We treat the true values of $f_1$ and $f_2$ as unknown, and add randomness to $f_1$ and $f_2$ to generate a simulation environment, i.e., $g_1\left(x_1,x_2\right)=f_1\left(x_1,x_2\right)+\epsilon_1$, $g_2\left(x_1,x_2\right)=f_2\left(x_1,x_2\right)+\epsilon_2$. Here, $\epsilon_1\sim N\left(0,10^{2}\right)$ and $\epsilon_2\sim N\left(0,3^{2}\right)$. The goal is to find the global minimum of $f_1$ and $f_2$ through multiple samples of $g_1$ and $g_2$.

\begin{figure}[!htbp]
	\centering
	\begin{subfigure}
		\centering
		\includegraphics[scale=0.52]{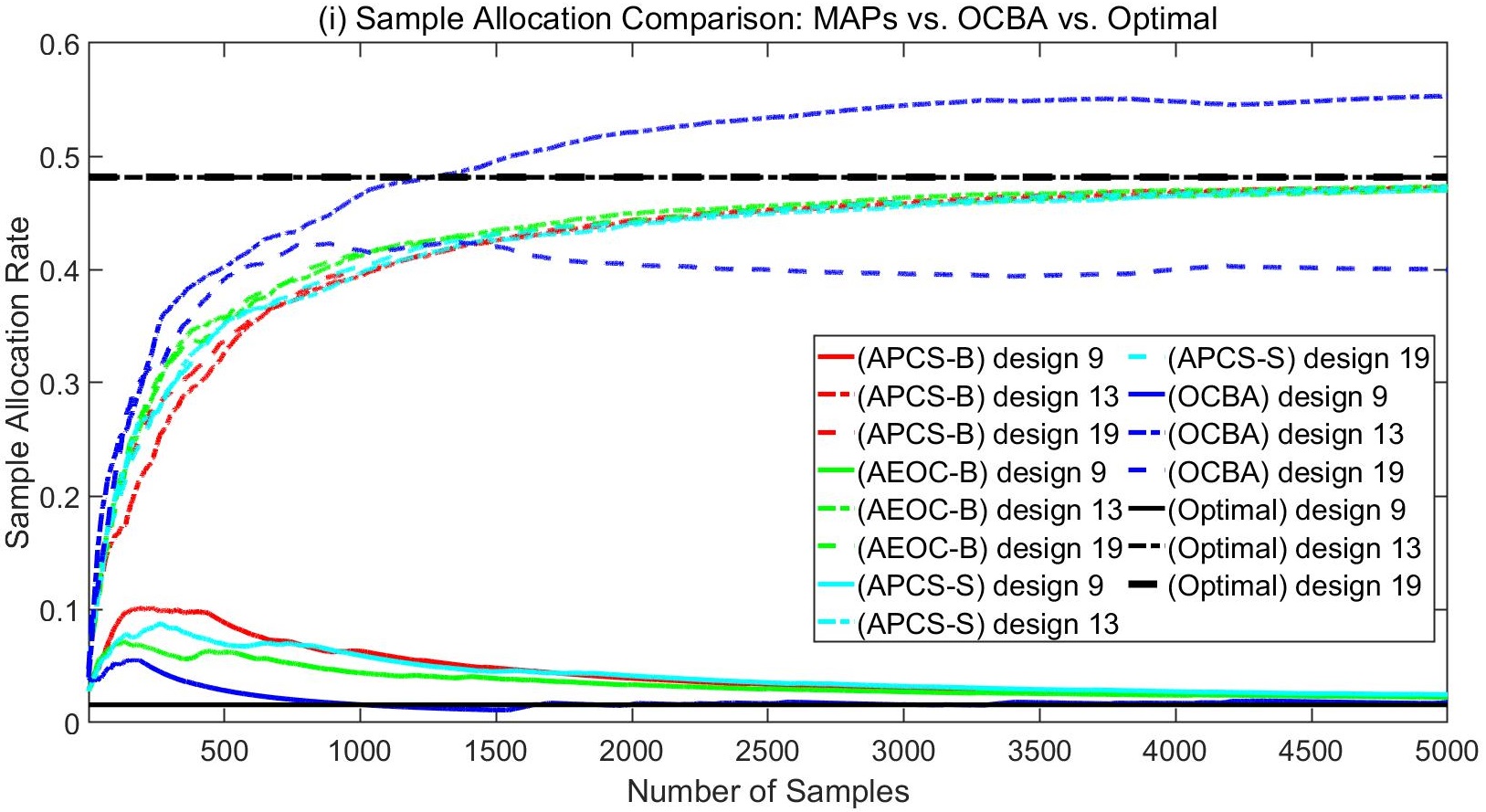}
	\end{subfigure}
	\vspace*{-3mm}
	\begin{subfigure}
		\centering
		\includegraphics[scale=0.52]{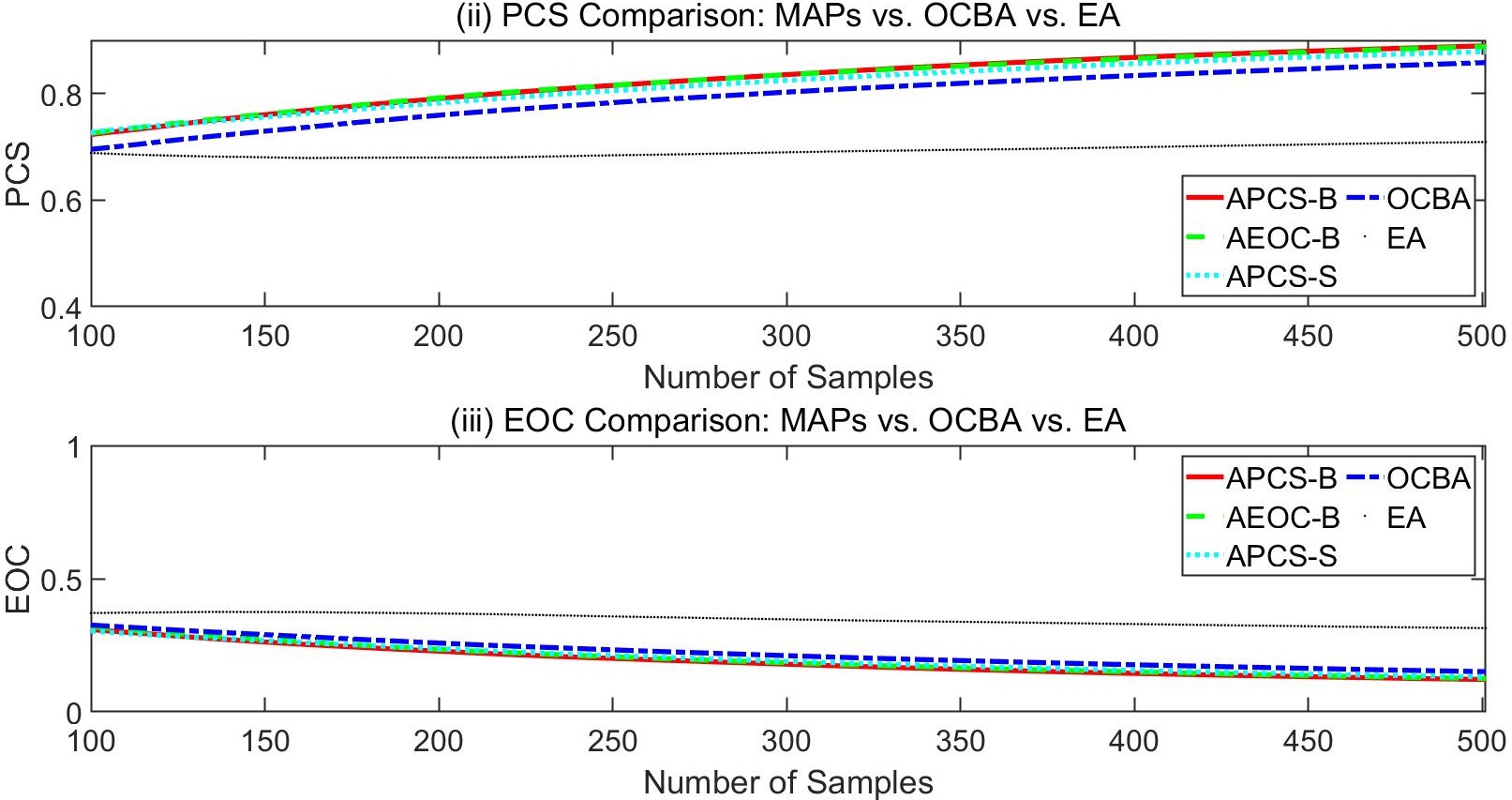}
	\end{subfigure}
	\caption{Rosenbrock Function (design 9: $\left(x_1,x_2\right)=(-1,1)$; design 13: $\left(x_1,x_2\right)=(0,0)$; design 19: $\left(x_1^{*},x_2^{*}\right)=(1,1))$.}
	\label{fig3}
\end{figure}

\begin{figure}[!htbp]
	\centering
	\begin{subfigure}
		\centering
		\includegraphics[scale=0.52]{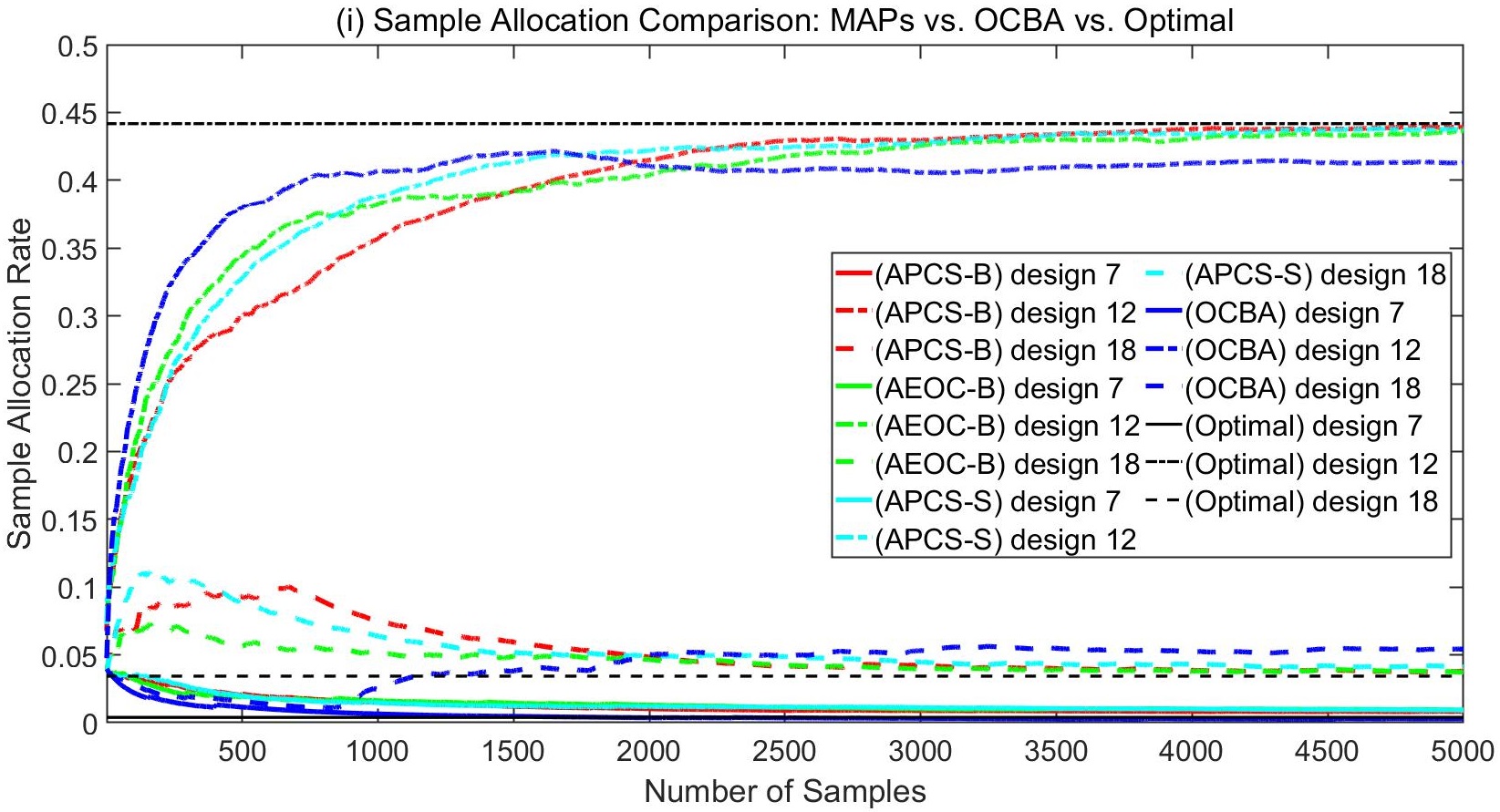}
	\end{subfigure}
	\vspace*{-3mm}
	\begin{subfigure}
		\centering
		\includegraphics[scale=0.52]{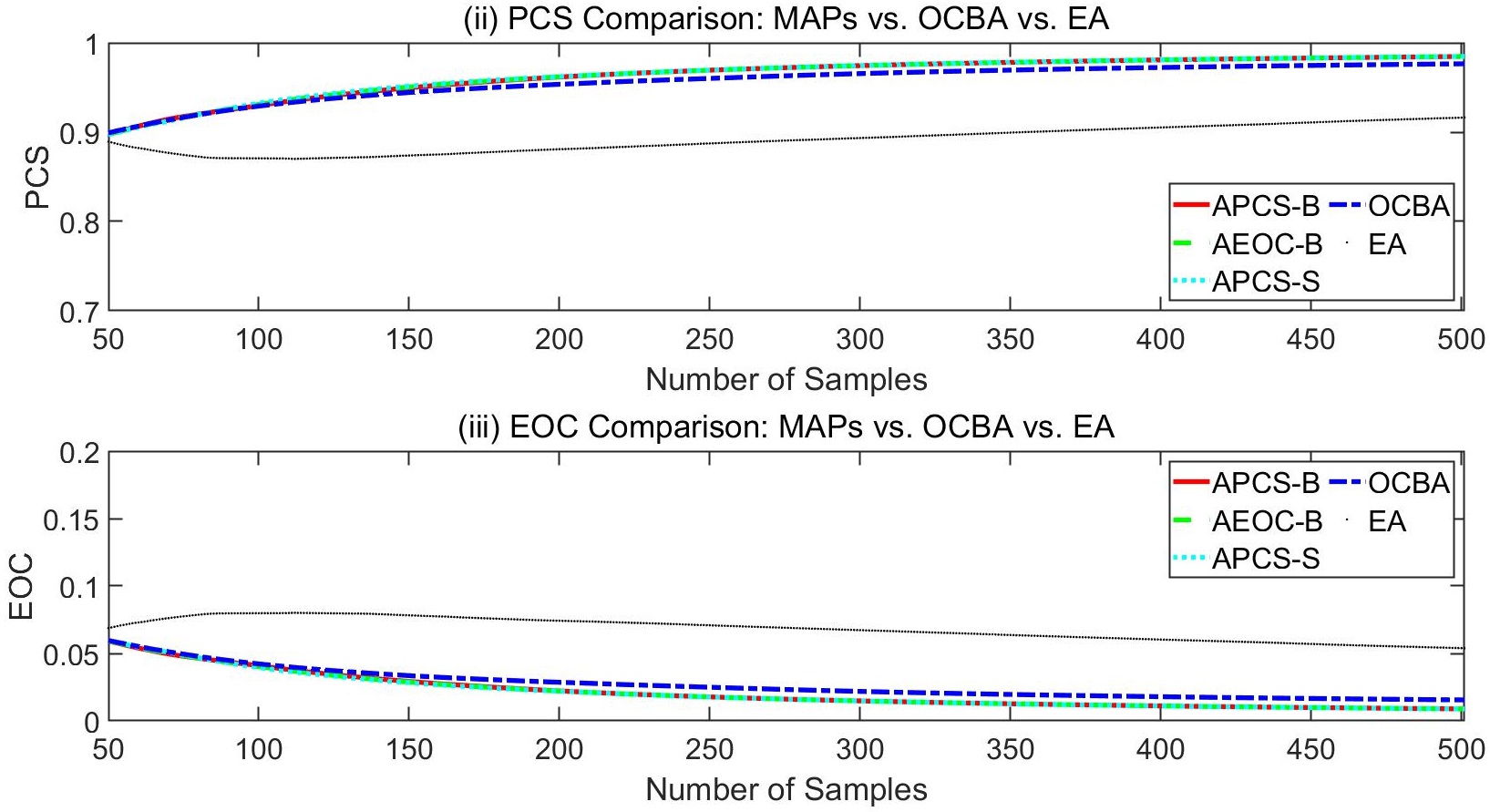}
	\end{subfigure}
	\caption{Goldstein-Price Function (design 7: $\left(x_1,x_2\right)=(-1,-1)$; design 12: $\left(x_1^{*},x_2^{*}\right)=(0,-1)$; design 18: $\left(x_1,x_2\right)=(1,0))$.}
	\label{fig4}
\end{figure}

The comparison results on the benchmark functions are reported in Figures \ref{fig3}-\ref{fig4}. Figure \ref{fig3}(i) shows that sample allocations of the MAPs on the three selected designs converge to the theoretically optimal values. In the first 2,000 samples, sample allocations of designs 13 and 19 from OCBA are closer to the optimal values than MAPs, while the MAPs exceed OCBA when the total number of samples is more than 2,000. For design 9, the sample allocation from OCBA is much closer to the optimal value compared to the MAPs. The three MAPs perform almost the same in the convergence of sample allocation. In Figures \ref{fig3}(ii) and \ref{fig3}(iii), PCS and EOC of the MAPs have slight advantages over those of OCBA and EA. The three MAPs have almost the same performance under PCS and EOC. EA performs the worst. The numerical results in Figure \ref{fig4} are similar to those in Figures \ref{fig1} and \ref{fig3}.

We further discuss some patterns observed in Figures \ref{fig1}-\ref{fig4}:
\begin{itemize}
	\item The MAPs outperform OCBA with respect to both PCS and EOC in general. A major reason is that OCBA tends to allocate more samples to the estimated best design in the early allocation stages compared to the MAPs, and causes some estimated non-best designs not sufficiently sampled. Then, if the initial estimate for the best design is not correct, a lot of samples are wasted on this incorrect design, and this mistake can only be fixed in later allocation stages when the estimated non-best designs start to receive additional samples. It undermines the performance of OCBA. In contrast, the MAPs allocate more samples to the estimated non-best designs in the early stages of the procedures. They can fix an incorrect initial estimate for the best design more quickly, and thus show better performance.
	\item Regarding sample allocations of designs, the MAPs have an explicit advantage over OCBA when the total number of samples is large enough. It is because OCBA follows (\ref{chen-ocba}), which are based on approximations (by assuming $N_{b}\gg N_i$, $i\ne b$) of the theoretically optimal conditions (\ref{ocba-nor}). Therefore for OCBA, the sample allocations of the designs converge to values that are slightly different from the optimal ones. The MAPs are directly built based on the optimality conditions (\ref{ocba-nor}), and are proved to converge to the optimal sample allocation.
	\item The performance of the three MAPs are similar. It suggests that the three MAPs studied in this research are not only asymptotically equivalent as proved, but also demonstrate similar finite-time performance in different structures of problems.
	\item The compared procedures have the same relative performance and trend under the measures of PCS and EOC when the sampling budget is large enough. This result is in line with the findings in \cite{gao2017} that PCS and EOC have the same rate function.
\end{itemize}

\subsection{Numerical Tests on Different Parameter Settings}

In this section, we assess the influence of $N_0$ and $\Delta$ on the performance of MAP and OCBA procedures using the Rosenbrock function. The Rosenbrock function follows the same settings as used in Section \ref{sec4.1}.

First, we study the influence of $N_0$ on the performance of the procedures. We set $\Delta=1$ and compare the performance of the procedures under $N_0=2,6,10$. Table \ref{tab1} shows the required sampling budgets for the MAP and OCBA procedures to attain ${\rm PCS}\geq 88\%$, ${\rm PCS}\geq 90\%$, ${\rm PCS}\geq 92\%$, ${\rm PCS}\geq 94\%$. We can see that under the setting of $N_0=2$, the MAPs use less sampling budgets than OCBA to achieve the desired values of PCS. Under the settings of $N_0=6,10$, AEOC-B and OCBA use less sampling budgets than APCS-B and APCS-S to achieve the desired values of PCS. In general, the procedures use the largest sampling budgets under $N_0=2$ and the smallest sampling budgets under $N_0=10$ to achieve the desired values of PCS.

\begin{table}[!ht]
	\centering
	\caption{\label{tab1}Comparison on the sampling budget among the MAPs and OCBA with $\Delta=1$ and different values of $N_0$'s.}
	\begin{tabular}{l|l|l|l|l|l}
		\hline
		%		\multicolumn{2}{l|}{} & \multicolumn{4}{c}{\rm PCS} \\ \hline
		\multicolumn{1}{c|}{$N_0$} & \multicolumn{1}{c|}{PCS} & \multicolumn{1}{c|}{APCS-B} & \multicolumn{1}{c|}{AEOC-B} & \multicolumn{1}{c|}{APCS-S} & \multicolumn{1}{c}{OCBA} \\ \hline
		\multirow{4}{*}{2} & \multicolumn{1}{c|}{88\%} & \multicolumn{1}{c|}{452} & \multicolumn{1}{c|}{460} & \multicolumn{1}{c|}{508} & \multicolumn{1}{c}{622} \\ \cline{2-6} 
		& \multicolumn{1}{c|}{90\%} & \multicolumn{1}{c|}{563} & \multicolumn{1}{c|}{571} & \multicolumn{1}{c|}{627} & \multicolumn{1}{c}{784} \\ \cline{2-6}
		& \multicolumn{1}{c|}{92\%} & \multicolumn{1}{c|}{718} & \multicolumn{1}{c|}{731} & \multicolumn{1}{c|}{792} & \multicolumn{1}{c}{1042} \\ \cline{2-6}
		& \multicolumn{1}{c|}{94\%} & \multicolumn{1}{c|}{975} & \multicolumn{1}{c|}{993} & \multicolumn{1}{c|}{1072} & \multicolumn{1}{c}{1481} \\ \hline
		\multirow{4}{*}{6} & \multicolumn{1}{c|}{88\%} & \multicolumn{1}{c|}{342} & \multicolumn{1}{c|}{274} & \multicolumn{1}{c|}{300} & \multicolumn{1}{c}{308} \\ \cline{2-6}
		& \multicolumn{1}{c|}{90\%} & \multicolumn{1}{c|}{438} & \multicolumn{1}{c|}{353} & \multicolumn{1}{c|}{387} & \multicolumn{1}{c}{393} \\ \cline{2-6}
		& \multicolumn{1}{c|}{92\%} & \multicolumn{1}{c|}{573} & \multicolumn{1}{c|}{468} & \multicolumn{1}{c|}{512} & \multicolumn{1}{c}{522} \\ \cline{2-6}
		& \multicolumn{1}{c|}{94\%} & \multicolumn{1}{c|}{779} & \multicolumn{1}{c|}{645} & \multicolumn{1}{c|}{703} & \multicolumn{1}{c}{721} \\  \hline
		\multirow{4}{*}{10} & \multicolumn{1}{c|}{88\%} & \multicolumn{1}{c|}{212} & \multicolumn{1}{c|}{199} & \multicolumn{1}{c|}{206} & \multicolumn{1}{c}{203} \\ \cline{2-6}
		& \multicolumn{1}{c|}{90\%} & \multicolumn{1}{c|}{297} & \multicolumn{1}{c|}{271} & \multicolumn{1}{c|}{290} & \multicolumn{1}{c}{278} \\ \cline{2-6}
		& \multicolumn{1}{c|}{92\%} & \multicolumn{1}{c|}{420} & \multicolumn{1}{c|}{370} & \multicolumn{1}{c|}{402} & \multicolumn{1}{c}{380} \\ \cline{2-6}
		& \multicolumn{1}{c|}{94\%} & \multicolumn{1}{c|}{592} & \multicolumn{1}{c|}{533} & \multicolumn{1}{c|}{574} & \multicolumn{1}{c}{534} \\  \hline
	\end{tabular}
\end{table}

Next, we study the influence of $\Delta$ on the performance of the procedures. We set $N_0=10$ and compare the performance of the procedures under $\Delta=1,5,10$. Table \ref{tab2} shows the required sampling budgets for the procedures to achieve ${\rm PCS}\geq 88\%$, ${\rm PCS}\geq 90\%$, ${\rm PCS}\geq 92\%$, ${\rm PCS}\geq 94\%$. Under the setting of $\Delta=1$, OCBA uses smaller sampling budgets than the MAPs to achieve the desired values of PCS. We can also see that under the settings of $\Delta=5,10$, OCBA uses larger sampling budgets than the MAPs. For the AEOC-B Procedure under $\Delta=1,5,10$, AEOC-B under $\Delta=1$ uses the smallest sampling budgets, and AEOC-B under $\Delta=10$ uses the largest sampling budgets to achieve the desired values of PCS. OCBA shows similar patterns to AEOC-B. For the APCS-B Procedure under $\Delta=1,5,10$ to achieve ${\rm PCS}\geq 88\%$, APCS-B under $\Delta=10$ uses the smallest sampling budget, and APCS-B under $\Delta=5$ uses the largest sampling budget. For the APCS-B Procedure under $\Delta=1,5,10$ to achieve ${\rm PCS}\geq 90\%$, APCS-B under $\Delta=1$ uses the smallest sampling budget, and APCS-B under $\Delta=5$ uses the largest sampling budget. For the APCS-B Procedure under $\Delta=1,5,10$ to achieve ${\rm PCS}\geq 92\%$ and ${\rm PCS}\geq 94\%$, APCS-B under $\Delta=1$ uses the smallest sampling budgets, and APCS-B under $\Delta=10$ uses the largest sampling budgets. The APCS-S Procedure shows similar patterns to the APCS-B Procedure.

\begin{table}[!ht]
	\centering
	\caption{\label{tab2}Comparison on the sampling budget among the MAPs and OCBA with $N_0=10$ and different values of $\Delta$'s.}
	\begin{tabular}{l|l|l|l|l|l}
		\hline
		%		\multicolumn{2}{l|}{} & \multicolumn{4}{c}{\rm PCS} \\ \hline
		\multicolumn{1}{c|}{$\Delta$} & \multicolumn{1}{c|}{PCS} & \multicolumn{1}{c|}{APCS-B} & \multicolumn{1}{c|}{AEOC-B} & \multicolumn{1}{c|}{APCS-S} & \multicolumn{1}{c}{OCBA} \\ \hline
		\multirow{4}{*}{1} & \multicolumn{1}{c|}{88\%} & \multicolumn{1}{c|}{212} & \multicolumn{1}{c|}{199} & \multicolumn{1}{c|}{206} & \multicolumn{1}{c}{203} \\ \cline{2-6} 
		& \multicolumn{1}{c|}{90\%} & \multicolumn{1}{c|}{297} & \multicolumn{1}{c|}{271} & \multicolumn{1}{c|}{290} & \multicolumn{1}{c}{278} \\ \cline{2-6}
		& \multicolumn{1}{c|}{92\%} & \multicolumn{1}{c|}{420} & \multicolumn{1}{c|}{370} & \multicolumn{1}{c|}{402} & \multicolumn{1}{c}{380} \\ \cline{2-6}
		& \multicolumn{1}{c|}{94\%} & \multicolumn{1}{c|}{592} & \multicolumn{1}{c|}{533} & \multicolumn{1}{c|}{574} & \multicolumn{1}{c}{534} \\  \hline
		\multirow{4}{*}{5} & \multicolumn{1}{c|}{88\%} & \multicolumn{1}{c|}{246} & \multicolumn{1}{c|}{208} & \multicolumn{1}{c|}{296} & \multicolumn{1}{c}{314} \\ \cline{2-6}
		& \multicolumn{1}{c|}{90\%} & \multicolumn{1}{c|}{356} & \multicolumn{1}{c|}{297} & \multicolumn{1}{c|}{406} & \multicolumn{1}{c}{428} \\ \cline{2-6}
		& \multicolumn{1}{c|}{92\%} & \multicolumn{1}{c|}{499} & \multicolumn{1}{c|}{423} & \multicolumn{1}{c|}{547} & \multicolumn{1}{c}{569} \\ \cline{2-6}
		& \multicolumn{1}{c|}{94\%} & \multicolumn{1}{c|}{712} & \multicolumn{1}{c|}{625} & \multicolumn{1}{c|}{760} & \multicolumn{1}{c}{791} \\  \hline
		\multirow{4}{*}{10} & \multicolumn{1}{c|}{88\%} & \multicolumn{1}{c|}{198} & \multicolumn{1}{c|}{233} & \multicolumn{1}{c|}{214} & \multicolumn{1}{c}{342} \\ \cline{2-6}
		& \multicolumn{1}{c|}{90\%} & \multicolumn{1}{c|}{347} & \multicolumn{1}{c|}{328} & \multicolumn{1}{c|}{346} & \multicolumn{1}{c}{486} \\ \cline{2-6}
		& \multicolumn{1}{c|}{92\%} & \multicolumn{1}{c|}{516} & \multicolumn{1}{c|}{463} & \multicolumn{1}{c|}{539} & \multicolumn{1}{c}{657} \\ \cline{2-6}
		& \multicolumn{1}{c|}{94\%} & \multicolumn{1}{c|}{748} & \multicolumn{1}{c|}{692} & \multicolumn{1}{c|}{801} & \multicolumn{1}{c}{918} \\  \hline
	\end{tabular}
\end{table}

We further discuss some patterns observed in Tables \ref{tab1} and \ref{tab2}:
\begin{itemize}
	\item When given a very small $N_0$ or a relatively large $\Delta$, the MAPs may perform better than OCBA under PCS. However, the advantages of the MAPs evaporate as $N_0$ gets larger or $\Delta$ gets smaller. 
	\item When given a relatively large budget, the MAP and OCBA procedures may perform better under a larger $N_0$ or a smaller $\Delta$.
	\item When given a small budget, APCS-B and APCS-S with a relatively large $\Delta$ might have better performance under PCS than those with a relatively small $\Delta$. It is because the estimates of the means of some designs may be inaccurate under a small budget, and a larger $\Delta$ provides the possibility to better estimate the means of the designs which could have caused the false estimation of the best design in some iterations. 
\end{itemize}

\section{Conclusions and Discussion}

R\&S, also known as ordinal optimization, is a well-established decision model for DEDS, and has been widely applied in manufacturing systems, traffic control, network reliability, and many other fields. For R\&S, consistency and sample allocation efficiency of the selection procedures have been two key questions under study. Consistency suggests correct selection for the best design with a large number of samples, and sample allocation efficiency evaluates the rates at which measures such as PCS and EOC converge to 1 and 0 as the sample allocation proceeds. In this research, we consider three MAPs, namely the APCS-B, AEOC-B and APCS-S Procedures, which were designed based on simple myopic heuristics. We conduct theoretical analysis on the three MAPs, and show that they are both consistent and efficient in sample allocation, where ``efficient'' is shown by satisfying the optimality conditions developed in \cite{chen2000a} and \cite{glynn2004}. It explains the excellent empirical performance of these procedures observed in the literature, and provides theoretical support for the application of them in practice. Moreover, it shows the
potential of procedures of this type, and indicates a possibly promising future research direction of developing superior MAPs using new approximations of the objective measures.

\bibliography{MAP_Ref}
\bibliographystyle{plain}

\newpage
\appendix
\onecolumn

\renewcommand\thesection{A.\arabic{section}}
\setcounter{equation}{0}
\renewcommand\theequation{A.\arabic{equation}}

\section{Proof of Lemma \ref{lem3}}\label{app_lem1}

We prove it by contradiction. Without loss of generality, suppose that $\lim_{n\to\infty}\left(\frac{\alpha_b^{\left(n\right)}}{\sigma_b}\right)^{2}-\sum_{i\ne b}\left(\frac{\alpha_i^{\left(n\right)}}{\sigma_i}\right)^{2}=0$ does not hold. It indicates that $\exists \epsilon_0>0$, $\forall T$, $\exists n_0>T$, $\left|\left(\frac{\alpha_b^{\left(n_0\right)}}{\sigma_b}\right)^{2}-\sum_{i\ne b}\left(\frac{\alpha_i^{\left(n_0\right)}}{\sigma_i}\right)^{2}\right|\geq\epsilon_0$. In addition, as $n\to\infty$, $N_i^{(n)}\to\infty$ for $\forall i$  leads to $L_i^{(n)}\to\infty$ for $\forall i$. Based on these conditions, we can find a subsequence $\left\{\alpha_i^{\left(n_t\right)}\Big{|}t=1,2,\dots,L_i^{\left(t\right)}\to\infty~\text{as}~t\to\infty,i=1,\dots,M\right\}$ satisfying that $\lim\inf_{t\to\infty}\left(\frac{\alpha_b^{\left(n_t\right)}}{\sigma_b}\right)^{2}-\sum_{i\ne b}\left(\frac{\alpha_i^{\left(n_t\right)}}{\sigma_i}\right)^{2}\geq\epsilon_0$ or $\lim\sup_{t\to\infty}\left(\frac{\alpha_b^{\left(n_t\right)}}{\sigma_b}\right)^{2}-\sum_{i\ne b}\left(\frac{\alpha_i^{\left(n_t\right)}}{\sigma_i}\right)^{2}\leq-\epsilon_0$. The subsequence $\left\{\alpha_i^{\left(n_t\right)}\Big{|}t=1,2,\dots,L_i^{\left(t\right)}\to\infty~\text{as}~t\to\infty,\right.$ $\left.i=1,\dots,M\rule{0em}{4mm}\right\}$ does not have any subsequence that satisfies (\ref{lem3-eq1}). It contradicts that each subsequence $\left\{\alpha_i^{\left(n_t\right)}\Big{|}t=1,2,\dots,L_i^{\left(t\right)}\to\infty~\text{as}~t\to\infty,\right.$ $\left.i=1,\dots,M\rule{0em}{4mm}\right\}$ has a further subsequence satisfying (\ref{lem3-eq1}). So the sequence $\left\{\alpha_i^{(n)}\Big{|}i=1,\dots,M,n=1,2,\dots\right\}$ satisfies (\ref{lem3-eq1}).
\hfill $\square$

\section{Proof of Theorem \ref{thm-consistency}}\label{app_thm1}

We consider a probability space $\left(\Omega,\mathcal{F},\mathbb{P}\right)$ in which all the normally distributed random variables are well defined. There exists a measurable sample space $\tilde{\Omega}\subseteq\Omega$ such that $\mathbb{P}\left(\tilde{\Omega}\right)=1$. According to the Strong Law of Large Numbers, for any sample path $\omega\in\tilde{\Omega}$, $\hat{\mu}_i^{(n)}\to\mu_i$ if $N_i^{(n)}\to\infty$ as $n\to\infty$ for $\forall i$. In addition, for any $\omega\in\tilde{\Omega}$, $\mathbb{P}\left(\hat{\mu}_i^{(n)}\ne\hat{\mu}_j^{(n)}\right)=1$, $\forall i,j$, $i\ne j$, $\forall n$, because the normal distributions are non-degenerate. Without loss of generality, we consider a fixed sample path $\omega\in\tilde{\Omega}$. For notation simplicity, we omit the dependence of the terms on $\omega$ when there is no ambiguity. Define $A=\left\{i\Big{|}N_i^{\left(n\right)}\to \infty~\text{as}~n\to \infty\right\}$, $B=\Big\{i\Big{|}\exists\zeta_i<\infty,\forall n>\zeta_i,N_i^{\left(n\right)}=N_i<\infty\Big\}$. $A\cup B=\left\{1,\dots,M\right\}$, $A\cap B=\emptyset$, $A\ne \emptyset$. To prove that $\lim_{n\to\infty}\hat{b}^{(n)}=b$, it is sufficient to show that $B=\emptyset$. Suppose that $B\ne \emptyset$. For $\forall i\in B$, $\forall n>\zeta_i$, $\hat{\mu}_{i}^{\left(n\right)}=\hat{\mu}_{i}$, $\hat{\sigma}_{i}^{\left(n\right)}=\hat{\sigma}_{i}$, $\mathbb{P}\left(\hat{\mu}_{i}\ne\mu_i\right)=1$. For $\forall i\in A$, $\hat{\mu}_{i}^{\left(n\right)}\to\mu_{i}$, $\hat{\sigma}_{i}^{\left(n\right)}\to\sigma_{i}$ as $n\to\infty$. Denote by $\hat{b}=\lim_{n\to \infty}\hat{b}^{\left(n\right)}$ $=\arg\max_i\left\{\mu_i\big{|}i\in A\right\}\bigcup\left\{\hat{\mu}_i\big{|}i\in B\right\}$. We first show the proof of Theorem \ref{thm-consistency} for the APCS-B Procedure.  The proof is divided into two stages:

\textbf{Stage 1:} Prove by contradiction that $\hat{b}\in A$. The proof is divided into three stages. Suppose that $\hat{b}\in B$.

\textbf{Stage 1(i):} Prove that $\hat{b}\in\arg\min_{i\in B}\left\{N_i\right\}$. Suppose that $\exists i_0\in B$, $N_{i_0}<N_{\hat{b}}$. For  $\text{APCS-B}^{(n),i_0}-\text{APCS-B}^{(n)}$, as $n\to\infty$, $s_{i_0,\hat{b}^{(n)}}^{(n)}\to s_{i_0,\hat{b}}^{(\infty)}$, $d_{i_0,\hat{b}^{(n)}}^{(n)}\to d_{i_0,\hat{b}}^{(\infty)}$, $\nu_{i_0,\hat{b}^{(n)}}^{(n)}\to \nu_{i_0,\hat{b}}^{(\infty)}$, $\tilde{s}_{i_0,\hat{b}^{(n)}}^{(n),i_0}\to \tilde{s}_{i_0,\hat{b}}^{(\infty),i_0}$, $\tilde{d}_{i_0,\hat{b}^{(n)}}^{(n),i_0}\to \tilde{d}_{i_0,\hat{b}}^{(\infty),i_0}$, $\tilde{\nu}_{i_0,\hat{b}^{(n)}}^{(n),i_0}\to \tilde{\nu}_{i_0,\hat{b}}^{(\infty),i_0}$, where $s_{i_0,\hat{b}}^{(\infty)}=\frac{\hat{\sigma}_{i_0}^{2}}{N_{i_0}}+\frac{\hat{\sigma}_{\hat{b}}^{2}}{N_{\hat{b}}}$,  $\nu_{i_0,\hat{b}}^{(\infty)}=\frac{\left(s_{i_0,\hat{b}}^{(\infty)}\right)^{2}}{\frac{\hat{\sigma}_{i_0}^{4}}{N_{i_0}^{2}\left(N_{i_0}-1\right)}+\frac{\hat{\sigma}_{\hat{b}}^{4}}{N_{\hat{b}}^{2}\left(N_{\hat{b}}-1\right)}}$,   $\tilde{\nu}_{i_0,\hat{b}}^{(\infty),i_0}=\frac{\left(\tilde{s}_{i_0,\hat{b}}^{(\infty),i_0}\right)^{2}}{\frac{\hat{\sigma}_{i_0}^{4}}{\left(N_{i_0}+1\right)^{2}N_{i_0}}+\frac{\hat{\sigma}_{\hat{b}}^{4}}{N_{\hat{b}}^{2}\left(N_{\hat{b}}-1\right)}}$, $\tilde{s}_{i_0,\hat{b}}^{(\infty),i_0}=\frac{\hat{\sigma}_{i_0}^{2}}{N_{i_0}+1}+\frac{\hat{\sigma}_{\hat{b}}^{2}}{N_{\hat{b}}}$, $d_{i_0,\hat{b}}^{(\infty)}=\frac{\hat{\mu}_{\hat{b}}-\hat{\mu}_{i_0}}{\sqrt{s_{i_0,\hat{b}}^{(\infty)}}}$, $\tilde{d}_{i_0,\hat{b}}^{(\infty),i_0}=\frac{\hat{\mu}_{\hat{b}}-\hat{\mu}_{i_0}}{\sqrt{\tilde{s}_{i_0,\hat{b}}^{(\infty),i_0}}}$. For $\nu_{i_0,\hat{b}}^{(\infty)}$, we have $\frac{\partial \nu_{i_0,\hat{b}}^{(\infty)}}{\partial N_{i_0}}>\frac{\frac{\hat{\sigma}_{i_0}^{4}}{N_{i_0}^{3}\left(N_{i_0}-1\right)^2}+\frac{\hat{\sigma}_{i_0}^{2}\hat{\sigma}_{\hat{b}}^{2}}{N_{i_0}^{2}N_{\hat{b}}}\left(\frac{3}{N_{i_0}\left(N_{i_0}-1\right)}-\frac{2}{N_{\hat{b}}\left(N_{\hat{b}}-1\right)}\right)}{\left(\frac{\hat{\sigma}_{i_0}^{2}}{N_{i_0}}+\frac{\hat{\sigma}_{\hat{b}}^{2}}{N_{\hat{b}}}\right)^{-1}\cdot\left(\frac{\left(\frac{\hat{\sigma}_{i_0}^{2}}{N_{i_0}}\right)^{2}}{N_{i_0}-1}+\frac{\left(\frac{\hat{\sigma}_{\hat{b}}^{2}}{N_{\hat{b}}}\right)^{2}}{N_{\hat{b}}-1}\right)^{2}}>0$ when $1<N_{i_0}<N_{\hat{b}}$. That is, when $1<N_{i_0}<N_{\hat{b}}$, $\nu_{i_0,\hat{b}}^{(\infty)}$ is monotone increasing with respect to $N_{i_0}$. So $\tilde{\nu}_{i_0,\hat{b}}^{(\infty),i_0}>\nu_{i_0,\hat{b}}^{(\infty)}$. Note that $\tilde{d}_{i_0,\hat{b}}^{(\infty),i_0}>d_{i_0,\hat{b}}^{(\infty)}$. Based on these conditions, $\lim_{n\to\infty}\text{APCS-B}^{(n),i_0}-\text{APCS-B}^{(n)}>0$. Meanwhile, for $j_0\in A$, $\nu_{j_0,\hat{b}^{(n)}}^{(n)}\to N_{\hat{b}}-1$, $d_{j_0,\hat{b}^{(n)}}^{(n)}\to\frac{\hat{\mu}_{\hat{b}}-\mu_{j_0}}{\hat{\sigma}_{\hat{b}}}\sqrt{N_{\hat{b}}}$, $\tilde{\nu}_{j_0,\hat{b}^{(n)}}^{(n),j_0}\to N_{\hat{b}}-1$, $\tilde{d}_{j_0,\hat{b}^{(n)}}^{(n),j_0}\to\frac{\hat{\mu}_{\hat{b}}-\mu_{j_0}}{\hat{\sigma}_{\hat{b}}}\sqrt{N_{\hat{b}}}$ as $n\to\infty$. Based on these conditions, $\lim_{n\to\infty}\text{APCS-B}^{(n),j_0}-\text{APCS-B}^{(n)}=0$. It indicates that $j_0$ receives at most finite samples as $n\to\infty$, which contradicts that $j_0\in A$. So $\hat{b}\in\arg\min_{i\in B}\left\{N_i\right\}$.

\textbf{Stage 1(ii):} Prove that $N_{\hat{b}}<N_i$ for $\forall i\in B$, $i\ne\hat{b}$. Suppose that $\exists i_0\in B$, $N_{i_0}=N_{\hat{b}}$. For $i_0\in B$, $j_0\in A$, by similar discussion in Stage 1(i), $\underset{n\to\infty}{\lim}\text{APCS-B}^{(n),i_0}-\text{APCS-B}^{(n)}>0$, $\underset{n\to\infty}{\lim}\text{APCS-B}^{(n),j_0}-\text{APCS-B}^{(n)}=0$. It contradicts that $j_0\in A$. So $N_{\hat{b}}<N_i$ for $\forall i\in B$, $i\ne\hat{b}$.

\textbf{Stage 1(iii):} Prove by contradiction that $\hat{b}\in A$. By similar discussion in Stage 1(i),  $\underset{n\to\infty}{\lim}\text{APCS-B}^{(n),\hat{b}}-\text{APCS-B}^{(n)}>0$, $\underset{n\to\infty}{\lim}\text{APCS-B}^{(n),j_0}-\text{APCS-B}^{(n)}=0$ for $\hat{b}\in B$, $j_0\in A$. It contradicts that $\hat{b}\in B$. So $\hat{b}\in A$.

\textbf{Stage 2:} Prove by contradiction that $B=\emptyset$. Suppose that $B\ne\emptyset$. By similar discussion as used in Stage 1(i), $\lim_{n\to\infty}\text{APCS-B}^{(n),\hat{b}}-\text{APCS-B}^{(n)}=0$, $\lim_{n\to\infty}\text{APCS-B}^{(n),i}-\text{APCS-B}^{(n)}>0$ for $\hat{b}\in A$, $\forall i\in B$. It contradicts that $i\in B$. So $B=\emptyset$. According to the Strong Law of Large Numbers, $\lim_{n\to\infty}\hat{b}^{(n)}=b$ for the APCS-B Procedure. The proofs of Theorem \ref{thm-consistency} for the AEOC-B and APCS-S Procedures have the same idea as above and are omitted for brevity.
\hfill $\square$

\section{Proof of Theorem \ref{thm-optsampalloc}}\label{app_thm2}

We first show the proof of Theorem \ref{thm-optsampalloc} for the APCS-B Procedure.  Before analyzing the theoretical performance of the APCS-B Procedure, we consider a modified version of $\text{APCS-B}^{(n)}$ and $\text{APCS-B}^{(n),j}$ for $\forall j$, $\text{mAPCS-B}^{\left(n\right)}=1-\sum_{i\ne \hat{b}^{(n)}}\Phi_{\underline{\nu}^{\left(n\right)}_{i,\hat{b}^{(n)}}}\left(-\underline{d}_{i,\hat{b}^{(n)}}^{\left(n\right)}\right)$, $\text{mAPCS-B}^{\left(n\right),j}=1-\sum_{i\ne \hat{b}^{(n)}}\Phi_{\underline{\tilde{\nu}}^{\left(n\right),j}_{i,\hat{b}^{(n)}}}\left(-\underline{\tilde{d}}_{i,\hat{b}^{(n)}}^{\left(n\right),j}\right)$, where  $\underline{d}_{i,\hat{b}^{(n)}}^{\left(n\right)}$, $\underline{\nu}^{\left(n\right)}_{i,\hat{b}^{(n)}}$,  $\underline{\tilde{d}}^{\left(n\right),j}_{i,\hat{b}^{(n)}}$, $\underline{\tilde{\nu}}^{\left(n\right),j}_{i,\hat{b}^{(n)}}$ are the same as the terms in (\ref{APCS-S}) and (\ref{tilde_d&s&nu_n}), except that the sample means and sample variances are replaced by their true values. We call the MAP using $\text{mAPCS-B}^{\left(n\right)}$ and $\text{mAPCS-B}^{\left(n\right),j}$ the mAPCS-B Procedure.
\begin{framed}[0.75\textwidth]
	\textbf{mAPCS-B Procedure}
	\begin{itemize}
		\item[1:] (lines 1-3 of APCS-B Procedure)
		\item[2:] \textbf{WHILE} $\sum_{i=1}^{M}N_i^{(n)}<N$ \textbf{DO}
		\item[3:] Update $\hat{\mu}_i^{(n)}$ for each $i$, and $\hat{b}^{(n)}$.
		\item[4:] $I^{\left(n\right)}=\mathop{\arg\max}_i \left[\text{mAPCS-B}^{\left(n\right),i}-\text{mAPCS-B}^{\left(n\right)}\right]$.
		\item[5:] (lines 7-8 of APCS-B Procedure)
		\item[6:] \textbf{END WHILE}
		\item[7:] Select $\hat{b}^{\left(N\right)}$ as the estimated best design.
	\end{itemize}
\end{framed}
This modification reduces some uncertainty in the allocation procedure (line 4 of the mAPCS-B Procedure) and makes the theoretical analysis more tractable. We fix a sample path $\omega\in\tilde{\Omega}$ and omit $\omega$ for notation simplicity. The proof is divided into three stages.

\textbf{Stage 1:} Prove that the mAPCS-B Procedure is consistent. The proof can be presented by similar discussion in Theorem \ref{thm-consistency} and thus omitted for brevity.

\textbf{Stage 2:} Prove that  $\left\{\underline{\alpha}_i^{(n)}\Big{|}i=1,\dots,M\right\}$ from the mAPCS-B Procedure satisfies (\ref{thm-opteqs1}) and (\ref{thm-opteqs2}). Based on Lemma \ref{lem3}, a sufficient condition of $\left\{\underline{\alpha}_i^{(n)}\Big{|}i=1,\dots,M\right\}$ satisfying (\ref{thm-opteqs1}) and (\ref{thm-opteqs2}) is that each subsequence $\left\{\underline{\alpha}_i^{\left(n_t\right)}\Big{|}i=1,\dots,M\right\}$ satisfying $L_i^{(t)}\to\infty$ as $t\to\infty$ for $\forall i$ has a convergent subsequence whose convergence point satisfies (\ref{ocba-nor}). We prove that the sufficient condition holds for $\left\{\underline{\alpha}_i^{(n)}\Big{|}i=1,\dots,M\right\}$. Based on Bolzano-Weierstrass theorem,  $\left\{\underline{\alpha}_i^{\left(n_t\right)}\Big{|}i=1,\dots,M\right\}$ satisfying $L_i^{(t)}\to\infty$ as $t\to\infty$ has convergent subsequences. Without loss of generality, denote by $\left\{\underline{\alpha}_i^{\left(n_{t_q}\right)}\Big{|}L_i^{(q)}\to\infty~\text{as}~q\to\infty,i=1,\dots,M\right\}$ any convergent subseuqence of $\left\{\underline{\alpha}_i^{\left(n_t\right)}\Big{|}i=1,\dots,M\right\}$ satisfying $L_i^{(t)}\to\infty$ as $t\to\infty$. Let $\left\{\underline{\alpha}_i\big{|}i=1,\dots,M\right\}$ be the convergence point of $\left\{\underline{\alpha}_i^{\left(n_{t_q}\right)}\Big{|}L_i^{(q)}\to\infty~\text{as}~q\to\infty,\forall i\right\}$. According to the Strong Law of Large Numbers, $\forall\epsilon>0$, $\exists\tilde{\zeta}$, $\forall n>\tilde{\zeta}$, $\left|\hat{\mu}_i^{(n)}-\mu_i\right|<\epsilon$ for $\forall i$. If we specify $\epsilon_0<\frac{1}{2}\min\left\{\left|\mu_i-\mu_j\right|\big{|}i,j\in\left\{1,\dots,M\right\},i\ne j\right\}$, $\exists\tilde{\zeta}_0$, $\forall n>\tilde{\zeta}_0$, $\hat{b}^{(n)}=b$. The proof of $\left\{\underline{\alpha}_i\big{|}i=1,\dots,M\right\}$ satisfying (\ref{ocba-nor}) is divided into five stages and is shown below.

\textbf{Stage 2(i):} Prove that $\underline{\alpha}_i\big{/}\underline{\alpha}_b<\infty$, $\forall i\ne b$. Suppose that $\exists i_0\ne b$, $\underline{\alpha}_{i_0}^{\left(n_{t_q}\right)}\big{/}\underline{\alpha}_{b}^{\left(n_{t_q}\right)}\to\infty$ as $q\to\infty$. Notice that $\lim_{q\to\infty}\left(\underline{\tilde{d}}_{i_0,b}^{\left(n_{t_q}\right),b}\right)^{2}-\left(\underline{d}_{i_0,b}^{\left(n_{t_q}\right)}\right)^{2}=\lim_{q\to\infty}\frac{\left(\mu_{i_0}-\mu_{b}\right)^{2}\sigma_{b}^{2}}{\left(\sigma_{i_0}^{2}\underline{\alpha}_{b}^{\left(n_{t_q}\right)}\Big{/}\underline{\alpha}_{i_0}^{\left(n_{t_q}\right)}+\sigma_{b}^{2}\right)^{2}}$. As $q\to\infty$, $\underline{\nu}_{i,b}^{\left(n_{t_q}\right)}\to\infty$, $\tilde{\underline{\nu}}_{i,b}^{\left(n_{t_q}\right),b}\to\infty$, $\forall i\ne b$, $\underline{\tilde{d}}_{i_0,b}^{\left(n_{t_q}\right),b}>\underline{d}_{i_0,b}^{\left(n_{t_q}\right)}$, $\underline{\tilde{d}}_{i,b}^{\left(n_{t_q}\right),b}\geq\underline{d}_{i,b}^{\left(n_{t_q}\right)}$, $\forall i\ne i_0,b$. So $\lim_{q\to\infty}\text{mAPCS-B}^{\left(n_{t_q}\right),b}-\text{mAPCS-B}^{\left(n_{t_q}\right)}>0$. In addition, for $\text{mAPCS-B}^{\left(n_{t_q}\right),i_0}-\text{mAPCS-B}^{\left(n_{t_q}\right)}$, as $q\to\infty$, $\underline{\nu}_{i,b}^{\left(n_{t_q}\right)}\to\infty$, $\tilde{\underline{\nu}}_{i,b}^{\left(n_{t_q}\right),i_0}\to\infty$, $\underline{\tilde{d}}_{i_0,b}^{\left(n_{t_q}\right),i_0}-\underline{d}_{i_0,b}^{\left(n_{t_q}\right)}\to 0$. So $\lim_{q\to\infty}\text{mAPCS-B}^{\left(n_{t_q}\right),i_0}-\text{mAPCS-B}^{\left(n_{t_q}\right)}=0$. It implies that $i_0$ receives at most finite samples in the limit of $q\to\infty$, which contradicts that $L_{i_0}^{(q)}\to\infty$ as $q\to\infty$. So  $\underline{\alpha}_i\big{/}\underline{\alpha}_b<\infty$, $\forall i\ne b$.

\textbf{Stage 2(ii):} Prove that $\lim_{q\to\infty}\underline{d}_{i,b}^{\left(n_{t_q}\right)}\Big{/}\underline{d}_{j,b}^{\left(n_{t_q}\right)}=1$, $\forall i,j\ne b$. Denote by $\underline{\tau}_i^{\left(n_{t_q}\right)}=\left(\underline{d}_{i,b}^{\left(n_{t_q}\right)}\right)^2\Big{/}n_{t_q}$, $\tilde{\underline{\tau}}_i^{\left(n_{t_q}\right),j}=\left(\underline{\tilde{d}}_{i,b}^{\left(n_{t_q}\right),j}\right)^2\Big{/}n_{t_q}$, $\forall i\ne b$, $\forall j\in\left\{1,\dots,M\right\}$. For $i_0\ne b$,
\begin{align*}
    &\lim_{q\to\infty}\text{mAPCS-B}^{\left(n_{t_q}\right),i_0}-\text{mAPCS-B}^{\left(n_{t_q}\right)}\\
    =&\underset{q\to\infty}{\lim}~\frac{\Phi_{\underline{\nu}_{i_0,b}^{\left(n_{t_q}\right)}}\left(-\sqrt{n_{t_q}\underline{\tau}_{i_0}^{\left(n_{t_q}\right)}}\right)-\Phi_{\underline{\nu}_{i_0,b}^{\left(n_{t_q}\right)}}\left(-\sqrt{n_{t_q}\tilde{\underline{\tau}}_{i_0}^{\left(n_{t_q}\right),i_0}}\right)}{n_{t_q}\tilde{\underline{\tau}}_{i_0}^{\left(n_{t_q}\right),i_0}-n_{t_q}\underline{\tau}_{i_0}^{\left(n_{t_q}\right)}}\cdot\frac{\tilde{\underline{\tau}}_{i_0}^{\left(n_{t_q}\right),i_0}-\underline{\tau}_{i_0}^{\left(n_{t_q}\right)}}{\underline{\alpha}_{i_0}^{\left(n_{t_q}\right)}+\frac{1}{n_{t_q}}-\underline{\alpha}_{i_0}^{\left(n_{t_q}\right)}}\\
    =&\underset{q\to\infty}{\lim}\kappa_{i_0}^{\left(n_{t_q}\right)}\phi_{\underline{\nu}_{i_0,b}^{\left(n_{t_q}\right)}}\left(-\underline{d}_{i_0,b}^{\left(n_{t_q}\right)}\right),
\end{align*}
where $\kappa_{i_0}^{\left(n_{t_q}\right)}=\frac{\left(\mu_b-\mu_{i_0}\right)^{2}\sigma_{i_0}^{2}}{2\underline{d}_{i_0,b}^{\left(n_{t_q}\right)}\left(\sigma_{i_0}^{2}+\sigma_b^{2}\underline{\alpha}_{i_0}^{\left(n_{t_q}\right)}\big{/}\underline{\alpha}_b^{\left(n_{t_q}\right)}\right)^{2}}$. Suppose $\exists\epsilon_0>0$, $\exists i_0,j_0\ne b$, $\underset{q\to\infty}{\lim\sup}~\underline{d}_{i_0,b}^{\left(n_{t_q}\right)}\Big{/}\underline{d}_{j_0,b}^{\left(n_{t_q}\right)}\geq\sqrt{1+\epsilon_0}$. There exists a subsequence $\left\{\underline{\alpha}_i^{\left(n_{t_{q_r}}\right)}\Big{|}i=1,\dots,M\right\}$ with $L_i^{(r)}\to\infty$ as $r\to\infty$ for $\forall i$ and  $\underline{d}_{i_0,b}^{\left(n_{t_{q_r}}\right)}\Big{/}\underline{d}_{j_0,b}^{\left(n_{t_{q_r}}\right)}\geq\sqrt{1+\epsilon_0}$ for large enough $n_{t_{q_r}}$. Notice that $\underline{\alpha}_i/\underline{\alpha}_b<\infty$ for $\forall i\ne b$.  Then, for $i_0,j_0\ne b$, $\exists\tilde{\zeta}''>\tilde{\zeta}'$, $\exists\underline{\delta}_1,\overline{\delta}_1,\delta_2>0$, $\forall n_{t_{q_r}}>\tilde{\zeta}''$, $\underline{\delta}_1\leq\kappa_{i_0}^{\left(n_{t_q}\right)}\big{/}\kappa_{j_0}^{\left(n_{t_q}\right)}\leq\overline{\delta}_1$, $\underline{\tau}_{j_0}^{\left(n_{t_{q_r}}\right)}\geq\delta_2$.
So
\begin{align*}
    &\underset{r\to\infty}{\lim}\frac{\text{mAPCS-B}^{\left(n_{t_{q_r}}\right),i_0}-\text{mAPCS-B}^{\left(n_{t_{q_r}}\right)}}{\text{mAPCS-B}^{\left(n_{t_{q_r}}\right),j_0}-\text{mAPCS-B}^{\left(n_{t_{q_r}}\right)}}\\
    \leq&\lim_{r\to\infty}\overline{\delta}_1\phi_{\underline{\nu}_{i_0,b}^{\left(n_{t_{q_r}}\right)}}\left(-\underline{d}_{i_0,b}^{\left(n_{t_{q_r}}\right)}\right)\Big{/}\phi_{\underline{\nu}_{j_0,b}^{\left(n_{t_{q_r}}\right)}}\left(-\underline{d}_{j_0,b}^{\left(n_{t_{q_r}}\right)}\right)\\
    =&\lim_{r\to\infty}\overline{\delta}_1\phi\left(-\underline{d}_{i_0,b}^{\left(n_{t_{q_r}}\right)}\right)\Big{/}\phi\left(-\underline{d}_{j_0,b}^{\left(n_{t_{q_r}}\right)}\right)\\
    \leq&\lim_{r\to\infty}\overline{\delta}_1\cdot\exp\left\{-\frac{\epsilon_0}{2}\delta_2 n_{t_{q_r}}\right\}=0
\end{align*}
where the second last equation holds because $\underline{\nu}_{i_0,b}^{\left(n_{t_{q_r}}\right)}\to\infty$, $\underline{\nu}_{j_0,b}^{\left(n_{t_{q_r}}\right)}\to\infty$ as $r\to\infty$, and $\underset{\nu\to\infty}{\lim}\frac{\phi_{\nu}(x)}{\phi(x)}=1$. It  indicates that $i_0$ receives at most finite samples in the limit of $r\to\infty$, which contradicts that $L_{i_0}^{(r)}\to\infty$ as $r\to\infty$. So $\lim\sup_{q\to\infty}\underline{d}_{i,b}^{\left(n_{t_q}\right)}\Big{/}\underline{d}_{j,b}^{\left(n_{t_q}\right)}\leq 1$, $\forall i,j\ne b$.
Next, assume that $\exists\epsilon_0>0$, $\exists i_0,j_0\ne b$, $\underset{q\to\infty}{\lim\inf}~\underline{d}_{i,b}^{\left(n_{t_q}\right)}\Big{/}\underline{d}_{j,b}^{\left(n_{t_q}\right)}\leq \sqrt{1-\epsilon_0}$. Following similar arguments as used above, $\frac{\text{mAPCS-B}^{\left(n_{t_{q_r}}\right),i_0}-\text{mAPCS-B}^{\left(n_{t_{q_r}}\right)}}{\text{mAPCS-B}^{\left(n_{t_{q_r}}\right),j_0}-\text{mAPCS-B}^{\left(n_{t_{q_r}}\right)}}\to\infty$ as $r\to\infty$. It contradicts that $L_{j_0}^{(r)}\to\infty$ as $r\to\infty$. So $\underset{q\to\infty}{\lim\inf}~\underline{d}_{i,b}^{\left(n_{t_q}\right)}\Big{/}\underline{d}_{j,b}^{\left(n_{t_q}\right)}\geq 1$, $\forall i,j\ne b$. Thus, $\underset{q\to\infty}{\lim}~\underline{d}_{i,b}^{\left(n_{t_q}\right)}\Big{/}\underline{d}_{j,b}^{\left(n_{t_q}\right)}=1$, $\forall i,j\ne b$.

\textbf{Stage 2(iii):} Prove that $\underline{\alpha}_i>0$ for $\forall i\in\left\{1,\dots,M\right\}$. First, we assume that $\forall i\ne b$, $\underline{\alpha}_{i}^{\left(n_{t_q}\right)}\big{/}\underline{\alpha}_{b}^{\left(n_{t_q}\right)}\to 0$ as $q\to\infty$. By similar discussion as used in Stage 2(i), $\lim_{q\to\infty}\text{mAPCS-B}^{\left(n_{t_q}\right),b}-\text{mAPCS-B}^{\left(n_{t_q}\right)}=0$, $\lim_{q\to\infty}\text{mAPCS-B}^{\left(n_{t_q}\right),i}-\text{mAPCS-B}^{\left(n_{t_q}\right)}>0$, $\forall i\ne b$. It contradicts that $L_{b}^{(q)}\to\infty$ as $q\to\infty$. So $\underline{\alpha}_i\big{/}\underline{\alpha}_b>0$ holds for some $i\ne b$. Furthermore, we assume that $\exists i_0\ne b$, $\underline{\alpha}_{i_0}^{\left(n_{t_q}\right)}\big{/}\underline{\alpha}_{b}^{\left(n_{t_q}\right)}\to 0$ as $q\to\infty$. Notice that $\exists j_0\ne b$, $\underline{\alpha}_{j_0}\big{/}\underline{\alpha}_b>0$, which indicates that $\underline{\alpha}_{i_0}^{\left(n_{t_q}\right)}\big{/}\underline{\alpha}_{j_0}^{\left(n_{t_q}\right)}\to 0$. Then, $\lim_{q\to\infty}\underline{d}_{i_0,b}^{\left(n_{t_q}\right)}\big{/}\underline{d}_{j_0,b}^{\left(n_{t_q}\right)}=0$, which contradicts that $\lim_{q\to\infty}\underline{d}_{i,b}^{\left(n_{t_q}\right)}\big{/}\underline{d}_{j,b}^{\left(n_{t_q}\right)}=1$, $\forall i,j\ne b$. That is, $\underline{\alpha}_i\big{/}\underline{\alpha}_b>0$, $\forall i\ne b$. So $\underline{\alpha}_i>0$ for $\forall i$.

\textbf{Stage 2(iv):} Prove by contradiction that  $\underset{q\to\infty}{\lim}\frac{\text{mAPCS-B}^{\left(n_{t_q}\right),i}-\text{mAPCS-B}^{\left(n_{t_q}\right)}}{\text{mAPCS-B}^{\left(n_{t_q}\right),j}-\text{mAPCS-B}^{\left(n_{t_q}\right)}}=1$, $\forall i,j$. Suppose that $\underset{q\to\infty}{\lim\sup}\frac{\text{mAPCS-B}^{\left(n_{t_q}\right),i_0}-\text{mAPCS-B}^{\left(n_{t_q}\right)}}{\text{mAPCS-B}^{\left(n_{t_q}\right),j_0}-\text{mAPCS-B}^{\left(n_{t_q}\right)}}>1$ for some $i_0,j_0$. Then, we can find a subsequence $\left\{\underline{\alpha}_i^{\left(n_{t_{q_r}}\right)}\Big{|}L_i^{(r)}\to\infty~\text{as}~r\to\infty,i=1,\dots,M\right\}$ satisfying\\ that $\underset{r\to\infty}{\lim}\frac{\text{mAPCS-B}^{\left(n_{t_{q_r}}\right),i_0}-\text{mAPCS-B}^{\left(n_{t_{q_r}}\right)}}{\text{mAPCS-B}^{\left(n_{t_{q_r}}\right),j_0}-\text{mAPCS-B}^{\left(n_{t_{q_r}}\right)}}>1$. It indicates that $j_0$ receives at most finite samples in the limit of $r\to\infty$, which contradicts that $L_{j_0}^{(r)}\to\infty$ as $r\to\infty$. So $\underset{q\to\infty}{\lim\sup}\frac{\text{mAPCS-B}^{\left(n_{t_q}\right),i}-\text{mAPCS-B}^{\left(n_{t_q}\right)}}{\text{mAPCS-B}^{\left(n_{t_q}\right),j}-\text{mAPCS-B}^{\left(n_{t_q}\right)}}\leq 1$ for $\forall i,j$. Similarly,  $\underset{q\to\infty}{\lim\inf}\frac{\text{mAPCS-B}^{\left(n_{t_q}\right),i}-\text{mAPCS-B}^{\left(n_{t_q}\right)}}{\text{mAPCS-B}^{\left(n_{t_q}\right),j}-\text{mAPCS-B}^{\left(n_{t_q}\right)}}\geq 1$, $\forall i,j$. So $\underset{q\to\infty}{\lim}\frac{\text{mAPCS-B}^{\left(n_{t_q}\right),i}-\text{mAPCS-B}^{\left(n_{t_q}\right)}}{\text{mAPCS-B}^{\left(n_{t_q}\right),j}-\text{mAPCS-B}^{\left(n_{t_q}\right)}}=1$, $\forall i,j$.

\textbf{Stage 2(v):} Prove that $\left(\underline{\alpha}_b\big{/}\sigma_b\right)^{2}-\sum_{i\ne b}\left(\underline{\alpha}_i\big{/}\sigma_i\right)^{2}=0$. Denote by $\underline{\tau}_i=\frac{\left(\mu_b-\mu_i\right)^{2}}{\frac{\sigma_i^{2}}{\underline{\alpha}_i}+\frac{\sigma_b^{2}}{\underline{\alpha}_b}}=\lim_{q\to\infty}\underline{\tau}_i^{\left(n_{t_q}\right)}$, $\forall i\ne b$. For $\forall j\ne b$,
\begin{align*}
    1=&\lim_{q\to\infty}\frac{\text{mAPCS-B}^{\left(n_{t_q}\right),b}-\text{mAPCS-B}^{\left(n_{t_q}\right)}}{\text{mAPCS-B}^{\left(n_{t_q}\right),j}-\text{mAPCS-B}^{\left(n_{t_q}\right)}}\\
    =&\lim_{q\to\infty}\sum_{i\ne b}\frac{\Phi_{\underline{\nu}_{i,b}^{\left(n_{t_q}\right)}}\left(-\underline{d}_{i,b}^{\left(n_{t_q}\right)}\right)-\Phi_{\underline{\tilde{\nu}}_{i,b}^{\left(n_{t_q}\right),b}}\left(-\tilde{\underline{d}}_{i,b}^{\left(n_{t_q}\right),b}\right)}{\Phi_{\underline{\nu}_{i,b}^{\left(n_{t_q}\right)}}\left(-\underline{d}_{i,b}^{\left(n_{t_q}\right)}\right)-\Phi_{\underline{\tilde{\nu}}_{i,b}^{\left(n_{t_q}\right),i}}\left(-\tilde{\underline{d}}_{i,b}^{\left(n_{t_q}\right),i}\right)}\\
    =&\underset{q\to\infty}{\lim}\underset{i\ne b}{\sum}\frac{\Phi_{\underline{\nu}_{i,b}^{\left(n_{t_q}\right)}}\left(-\underline{d}_{i,b}^{\left(n_{t_q}\right)}\right)-\Phi_{\underline{\nu}_{i,b}^{\left(n_{t_q}\right)}}\left(-\tilde{\underline{d}}_{i,b}^{\left(n_{t_q}\right),b}\right)}{\Phi_{\underline{\nu}_{i,b}^{\left(n_{t_q}\right)}}\left(-\underline{d}_{i,b}^{\left(n_{t_q}\right)}\right)-\Phi_{\underline{\nu}_{i,b}^{\left(n_{t_q}\right)}}\left(-\tilde{\underline{d}}_{i,b}^{\left(n_{t_q}\right),i}\right)}\\
    =&\underset{q\to\infty}{\lim}\underset{i\ne b}{\sum}\frac{\frac{\tilde{\underline{\tau}}_{i}^{\left(n_{t_q}\right),b}-\underline{\tau}_{i}^{\left(n_{t_q}\right)}}{\underline{\alpha}_{b}^{\left(n_{t_q}\right)}+\frac{1}{n_{t_q}}-\underline{\alpha}_{b}^{\left(n_{t_q}\right)}}\frac{\Phi_{\underline{\nu}_{i,b}^{\left(n_{t_q}\right)}}\left(-\tilde{\underline{d}}_{i,b}^{\left(n_{t_q}\right),b}\right)-\Phi_{\underline{\nu}_{i,b}^{\left(n_{t_q}\right)}}\left(-\underline{d}_{i,b}^{\left(n_{t_q}\right)}\right)}{\tilde{\underline{\tau}}_{i}^{\left(n_{t_q}\right),b}-\underline{\tau}_{i}^{\left(n_{t_q}\right)}}}{\frac{\tilde{\underline{\tau}}_{i}^{\left(n_{t_q}\right),i}-\underline{\tau}_{i}^{\left(n_{t_q}\right)}}{\underline{\alpha}_{i}^{\left(n_{t_q}\right)}+\frac{1}{n_{t_q}}-\underline{\alpha}_{i}^{\left(n_{t_q}\right)}}\frac{\Phi_{\underline{\nu}_{i,b}^{\left(n_{t_q}\right)}}\left(-\tilde{\underline{d}}_{i,b}^{\left(n_{t_q}\right),i}\right)-\Phi_{\underline{\nu}_{i,b}^{\left(n_{t_q}\right)}}\left(-\underline{d}_{i,b}^{\left(n_{t_q}\right)}\right)}{\tilde{\underline{\tau}}_{i}^{\left(n_{t_q}\right),i}-\underline{\tau}_{i}^{\left(n_{t_q}\right)}}}\\
    =&\underset{q\to\infty}{\lim}\underset{i\ne b}{\sum}\frac{\frac{\tilde{\underline{\tau}}_{i}^{\left(n_{t_q}\right),b}-\underline{\tau}_{i}^{\left(n_{t_q}\right)}}{\underline{\alpha}_{b}^{\left(n_{t_q}\right)}+\frac{1}{n_{t_q}}-\underline{\alpha}_{b}^{\left(n_{t_q}\right)}}\frac{\sqrt{n_{t_q}}}{2\sqrt{\underline{\tau}_{i}^{\left(n_{t_q}\right)}}}\phi_{\underline{\nu}_{i,b}^{\left(n_{t_q}\right)}}\left(-\underline{d}_{i,b}^{\left(n_{t_q}\right)}\right)}{\frac{\tilde{\underline{\tau}}_{i}^{\left(n_{t_q}\right),i}-\underline{\tau}_{i}^{\left(n_{t_q}\right)}}{\underline{\alpha}_{i}^{\left(n_{t_q}\right)}+\frac{1}{n_{t_q}}-\underline{\alpha}_{i}^{\left(n_{t_q}\right)}}\frac{\sqrt{n_{t_q}}}{2\sqrt{\underline{\tau}_{i}^{\left(n_{t_q}\right)}}}\phi_{\underline{\nu}_{i,b}^{\left(n_{t_q}\right)}}\left(-\underline{d}_{i,b}^{\left(n_{t_q}\right)}\right)}\\
    =&\underset{q\to\infty}{\lim}\underset{i\ne b}{\sum}\frac{\frac{\tilde{\underline{\tau}}_{i}^{\left(n_{t_q}\right),b}-\underline{\tau}_{i}^{\left(n_{t_q}\right)}}{\underline{\alpha}_{b}^{\left(n_{t_q}\right)}+\frac{1}{n_{t_q}}-\underline{\alpha}_{b}^{\left(n_{t_q}\right)}}}{\frac{\tilde{\underline{\tau}}_{i}^{\left(n_{t_q}\right),i}-\underline{\tau}_{i}^{\left(n_{t_q}\right)}}{\underline{\alpha}_{i}^{\left(n_{t_q}\right)}+\frac{1}{n_{t_q}}-\underline{\alpha}_{i}^{\left(n_{t_q}\right)}}}=\underset{i\ne b}{\sum}\frac{\frac{\partial\underline{\tau}_i}{\partial \underline{\alpha}_b}}{\frac{\partial\underline{\tau}_i}{\partial\underline{\alpha}_i}}=\frac{\underset{i\ne b}{\sum}\left(\frac{\underline{\alpha}_i}{\sigma_i}\right)^{2}}{\left(\frac{\underline{\alpha}_b}{\sigma_b}\right)^{2}}.
\end{align*}
The first and second equations hold based on the conclusion of Stage 2(iv). The third equation holds because 
%$\underset{q\to\infty}{\lim}\Phi_{\underline{\nu}_{i,b}^{\left(n_{t_q}\right)}}\left(-\underline{\tilde{d}}_{i,b}^{\left(n_{t_q}\right),j}\right)-\Phi_{\underline{\tilde{\nu}}_{i,b}^{\left(n_{t_q}\right),j}}\left(-\underline{\tilde{d}}_{i,b}^{\left(n_{t_q}\right),j}\right)=0$ while $\underset{q\to\infty}{\lim}\Phi_{\underline{\nu}_{i,b}^{\left(n_{t_q}\right)}}\left(-\underline{d}_{i,b}^{\left(n_{t_q}\right)}\right)-\Phi_{\underline{\nu}_{i,b}^{\left(n_{t_q}\right)}}\left(-\underline{\tilde{d}}_{i,b}^{\left(n_{t_q}\right),j}\right)>0$ 
$\underset{q\to\infty}{\lim}\frac{\Phi_{\underline{\nu}_{i,b}^{\left(n_{t_q}\right)}}\left(-\underline{\tilde{d}}_{i,b}^{\left(n_{t_q}\right),j}\right)-\Phi_{\underline{\tilde{\nu}}_{i,b}^{\left(n_{t_q}\right),j}}\left(-\underline{\tilde{d}}_{i,b}^{\left(n_{t_q}\right),j}\right)}{\Phi_{\underline{\nu}_{i,b}^{\left(n_{t_q}\right)}}\left(-\underline{d}_{i,b}^{\left(n_{t_q}\right)}\right)-\Phi_{\underline{\nu}_{i,b}^{\left(n_{t_q}\right)}}\left(-\underline{\tilde{d}}_{i,b}^{\left(n_{t_q}\right),j}\right)}=0$ resulting from $\lim_{\nu\to\infty}\frac{\phi_{\nu}(x)}{\phi(x)}=1$ and $\underline{\tilde{d}}_{i,b}^{\left(n_{t_q}\right),j}>\underline{d}_{i,b}^{\left(n_{t_q}\right)}$ as $q\to\infty$, $\forall i\ne b$, $\forall j$. The last four equations hold based on the definition of differentiation, the chain rule of differentiation, and $\lim_{q\to\infty}\underline{\alpha}_i^{\left(n_{t_q}\right)}=\underline{\alpha}_i$ for $\forall i$. Notice that $\sum_{i\ne b}\left(\underline{\alpha}_i\big{/}\sigma_i\right)^{2}\big{/}\left(\underline{\alpha}_b\big{/}\sigma_b\right)^{2}=1$ and $0<\underline{\alpha}_i<1$, $\forall i$. So $\left(\underline{\alpha}_b\big{/}\sigma_b\right)^{2}-\sum_{i\ne b}\left(\underline{\alpha}_i\big{/}\sigma_i\right)^{2}=0$. Based on the discussion in Stages 2(i)-2(v), we show that $\left\{\underline{\alpha}_i\big{|}i=1,\dots,M\right\}$ satisfies the optimality conditions (\ref{ocba-nor}). Thus, $\left\{\underline{\alpha}_i^{(n)}\Big{|}i=1,\dots,M\right\}$ satisfies (\ref{thm-opteqs1}) and (\ref{thm-opteqs2}).

\textbf{Stage 3:} Prove that $\left\{\alpha_i^{(n)}\Big{|}i=1,\dots,M\right\}$ from the APCS-B Procedure satisfies (\ref{thm-opteqs1}) and (\ref{thm-opteqs2}). Denote by $\tau_i^{\left(n\right)}=\frac{\left(\mu_b-\mu_i\right)^{2}}{\frac{\sigma_i^{2}}{\alpha_i^{\left(n\right)}}+\frac{\sigma_b^{2}}{\alpha_b^{\left(n\right)}}}$, $\tilde{\tau}_{i}^{\left(n\right),j}=\frac{\left(\mu_{b}-\mu_{i}\right)^{2}}{\frac{\sigma_{i}^{2}}{\alpha^{\left(n\right)}_{i}+\frac{\mathbbm{1}\left\{j=i\right\}}{n}}+\frac{\sigma_{b}^{2}}{\alpha_b^{\left(n\right)}+\frac{\mathbbm{1}\left\{j=b\right\}}{n}}}$, $\forall i\ne b$, $\forall j$.
For small enough $\epsilon>0$, $\exists\bar{\zeta}$, $\forall n>\bar{\zeta}$, one of the following two cases will occur for $\forall i\ne b$, $\forall j$:
\begin{align}
&0<\frac{\left(d_{i,b}^{(n)}\right)^{2}}{n}-\tau_i^{(n)}\leq\frac{\left(\tilde{d}_{i,b}^{(n),j}\right)^{2}}{n}-\tilde{\tau}_i^{(n),j}<\epsilon,\label{thm2-apcsb-d-ub}\\
&0<\tau_i^{(n)}-\frac{\left(d_{i,b}^{(n)}\right)^{2}}{n}\leq\tilde{\tau}_i^{(n),j}-\frac{\left(\tilde{d}_{i,b}^{(n),j}\right)^{2}}{n}<\epsilon.\label{thm2-apcsb-d-lb}
\end{align}	
In the case of (\ref{thm2-apcsb-d-ub}), $d_{i,b}^{\left(n\right)}\leq\sqrt{n\left(\tau_i^{(n)}+\epsilon\right)}$, $\sqrt{n\left(\tau_i^{(n)}+\epsilon\right)}-d_{i,b}^{\left(n\right)}\geq\sqrt{n\left(\tilde{\tau}_i^{(n),i}+\epsilon\right)}-\tilde{d}_{i,b}^{\left(n\right),i}$. Based on these conditions,
\begin{align*}
    \text{APCS-B}^{\left(n\right),i}-\text{APCS-B}^{\left(n\right)}\geq\Phi_{\nu_{i,b}^{\left(n\right)}}\left(-\sqrt{n\left(\tau_i^{(n)}+\epsilon\right)}\right)-\Phi_{\tilde{\nu}_{i,b}^{\left(n\right),i}}\left(-\sqrt{n\left(\tilde{\tau}_i^{(n),i}+\epsilon\right)}\right)\triangleq\underline{v}_{i}^{(n)}.
\end{align*}
It also holds that
\begin{align*}
    \text{APCS-B}^{\left(n\right),b}-\text{APCS-B}^{\left(n\right)}\geq\underset{i\ne b}{\sum}\left[\Phi_{\nu_{i,b}^{\left(n\right)}}\left(-\sqrt{n\left(\tau_i^{(n)}+\epsilon\right)}\right)-\Phi_{\tilde{\nu}_{i,b}^{\left(n\right),b}}\left(-\sqrt{n\left(\tilde{\tau}_i^{(n),b}+\epsilon\right)}\right)\right]\triangleq\underline{v}_{b}^{(n)}.
\end{align*}
In the case of (\ref{thm2-apcsb-d-lb}), it holds that for $i\ne b$,
\begin{align*}
    \text{APCS-B}^{\left(n\right),i}-\text{APCS-B}^{\left(n\right)}\leq\Phi_{\nu_{i,b}^{\left(n\right)}}\left(-\sqrt{n\left(\tau_i^{(n)}-\epsilon\right)}\right)-\Phi_{\tilde{\nu}_{i,b}^{\left(n\right),i}}\left(-\sqrt{n\left(\tilde{\tau}_i^{(n),i}-\epsilon\right)}\right)\triangleq\overline{v}_{i}^{(n)};
\end{align*}
in the meantime, 
\begin{align*}
    \text{APCS-B}^{\left(n\right),b}-\text{APCS-B}^{\left(n\right)}\leq\underset{i\ne b}{\sum}\left[\Phi_{\nu_{i,b}^{\left(n\right)}}\left(-\sqrt{n\left(\tau_i^{(n)}-\epsilon\right)}\right)-\Phi_{\tilde{\nu}_{i,b}^{\left(n\right),b}}\left(-\sqrt{n\left(\tilde{\tau}_i^{(n),b}-\epsilon\right)}\right)\right]\triangleq\overline{v}_{b}^{(n)}.
\end{align*}
Now we consider a hybrid procedure: When $n\leq\bar{\zeta}$, the sample allocation is generated based on the APCS-B Procedure. When $n>\bar{\zeta}$, the sample allocation is manipulated by a slightly adjusted version of mAPCS-B Procedure (called the \~{m}APCS-B Procedure). In the case of (\ref{thm2-apcsb-d-ub}), one always replaces $\text{mAPCS-B}^{\left(n\right),i}-\text{mAPCS-B}^{\left(n\right)}$ by $\underline{v}_{i}^{(n)}$ to evaluate  $i\ne b$, replaces $\text{mAPCS-B}^{\left(n\right),b}-\text{mAPCS-B}^{\left(n\right)}$ by $\underline{v}_{b}^{(n)}$ to evaluate $b$; in the case of (\ref{thm2-apcsb-d-lb}), one always replaces $\text{mAPCS-B}^{\left(n\right),i}-\text{mAPCS-B}^{\left(n\right)}$ by $\overline{v}_{i}^{(n)}$ to evaluate $i\ne b$, replaces $\text{mAPCS-B}^{\left(n\right),b}-\text{mAPCS-B}^{\left(n\right)}$ by $\overline{v}_{b}^{(n)}$ to evaluate $b$. For the \~{m}APCS-B Procedure, by similar discussion as used in Stage 2,  $\lim_{n\to\infty}\left|\tau_{i}^{(n)}-\tau_{j}^{(n)}\right|\leq 2\epsilon$ for $\forall i,j\ne b$, $\underset{n\to\infty}{\lim}\left|\left(\alpha_{b}^{\left(n\right)}\big{/}\sigma_{b}\right)^{2}-\sum_{i\ne b}\left(\alpha_{i}^{\left(n\right)}\big{/}\sigma_{i}\right)^{2}\right|\leq c\epsilon$ for some finite $c>0$. For $\forall i$, as $n>\bar{\zeta}$, this hybrid procedure always assigns design $i$ an improvement smaller than $\text{APCS-B}^{\left(n\right),i}-\text{APCS-B}^{\left(n\right)}$ in the case of (\ref{thm2-apcsb-d-ub}), and assigns design $i$ an improvement larger than $\text{APCS-B}^{\left(n\right),i}-\text{APCS-B}^{\left(n\right)}$ in the case of (\ref{thm2-apcsb-d-lb}). It implies that for $\forall i\in\left\{1,\dots,M\right\}$, the hybrid procedure samples design $i$ at least as often as the APCS-B Procedure in the case of (\ref{thm2-apcsb-d-ub}), and sample design $i$ at most as often as the APCS-B Procedure in the case of (\ref{thm2-apcsb-d-lb}). So the sample allocation from the APCS-B Procedure is lower-bounded by that from the \~{m}APCS-B Procedure in the case of (\ref{thm2-apcsb-d-ub}), and is upper-bounded by that from the \~{m}APCS-B Procedure in the case of (\ref{thm2-apcsb-d-lb}). Then, as $\epsilon\to 0$,  $\lim_{n\to\infty}\tau_{i}^{(n)}-\tau_{j}^{(n)}=0$ for $\forall i,j\ne b$, $\underset{n\to\infty}{\lim}\left(\alpha_{b}^{\left(n\right)}\big{/}\sigma_{b}\right)^{2}-\sum_{i\ne b}\left(\alpha_{i}^{\left(n\right)}\big{/}\sigma_{i}\right)^{2}=0$ for the APCS-B Procedure. Thus, the sample allocation $\left\{\alpha_i^{(n)}\Big{|}i=1,\dots,M\right\}$ from the APCS-B Procedure asymptotically satisfies (\ref{thm-opteqs1}) and (\ref{thm-opteqs2}). 

Next, we show the proof of Theorem 2 for the AEOC-B Procedure. Before analyzing the theoretical performance of the AEOC-B Procedure, we consider a modified version of $\text{AEOC-B}^{(n)}$ and $\text{AEOC-B}^{(n),j}$, given by
$\text{mAEOC-B}^{\left(n\right)}=\sum_{i\ne \hat{b}^{(n)}}\sqrt{\underline{s}_{i,\hat{b}^{(n)}}^{\left(n\right)}}\Psi_{\underline{\nu}^{\left(n\right)}_{i,\hat{b}^{(n)}}}\left(\underline{d}_{i,\hat{b}^{(n)}}^{\left(n\right)}\right)$,
$\text{mAEOC-B}^{\left(n\right),j}=\sum_{i\ne \hat{b}^{(n)}}\sqrt{\underline{\tilde{s}}_{i,\hat{b}^{(n)}}^{\left(n\right),j}}\Psi_{\underline{\tilde{\nu}}^{\left(n\right),j}_{i,\hat{b}^{(n)}}}\left(\underline{\tilde{d}}_{i,\hat{b}^{(n)}}^{\left(n\right),j}\right),\forall j,$ where terms $\underline{s}_{i,\hat{b}^{(n)}}^{\left(n\right)}$, $\underline{d}_{i,\hat{b}^{(n)}}^{\left(n\right)}$, $\underline{\nu}_{i,\hat{b}^{(n)}}^{\left(n\right)}$, $\underline{\tilde{s}}_{i,\hat{b}^{(n)}}^{\left(n\right),j}$, $\underline{\tilde{d}}_{i,\hat{b}^{(n)}}^{\left(n\right),j}$, $\underline{\tilde{\nu}}_{i,\hat{b}^{(n)}}^{\left(n\right),j}$ are the same as those defined in (\ref{AEOC-B}) and (\ref{tilde_d&s&nu_n}), except that the sample means and sample variances are replaced by their true values. We call the MAP using $\text{mAEOC-B}^{(n)}$ and $\text{mAEOC-B}^{(n),j}$ the mAEOC-B Procedure. We fix a sample path $\omega\in\tilde{\Omega}$ and omit $\omega$ for notation simplicity. The proof of Theorem \ref{thm-optsampalloc} for the AEOC-B Procedure is divided into three stages.

\textbf{Stage 1:} Prove that the mAEOC-B Procedure is consistent. The proof can be presented by similar discussion in Theorem 1 and thus omitted for brevity.

\textbf{Stage 2:} Prove that the sample allocation sequence $\left\{\underline{\alpha}_i^{(n)}\Big{|}i=1,\dots,M\right\}$ from the mAEOC-B Procedure satisfies (\ref{lem3-eq1}). It suffices to prove that each subsequence $\left\{\underline{\alpha}_i^{\left(n_t\right)}\Big{|}i=1,\dots,M,t=1,2,\dots\right\}$ satisfying $L_i^{(t)}\to\infty$ as $t\to\infty$ has a convergent subsequence whose convergence point satisfies the optimality conditions (\ref{ocba-nor}). Without loss of generality, denote by $\left\{\underline{\alpha}_i^{\left(n_{t_q}\right)}\Big{|}L_i^{(q)}\to\infty~\text{as}~q\to\infty,i=1,\dots,M\right\}$ any convergent subseuqence of $\left\{\underline{\alpha}_i^{\left(n_t\right)}\Big{|}i=1,\dots,M\right\}$ satisfying $L_i^{(t)}\to\infty~\text{as}~t\to\infty$. Let $\left\{\underline{\alpha}_i\big{|}i=1,\dots,M\right\}$ be the convergence point of the convergent subseuqence. The proof of $\left\{\underline{\alpha}_i\big{|}i=1,\dots,M\right\}$ satisfying (\ref{ocba-nor}) is divided into five stages and is shown as follows.

\textbf{Stage 2(i):} Prove by contradiction that $\underline{\alpha}_i/\underline{\alpha}_b<\infty$ for $\forall i\ne b$. Suppose that $\exists i_0\ne b$, $\underline{\alpha}_{i_0}^{\left(n_{t_q}\right)}\big{/}\underline{\alpha}_{b}^{\left(n_{t_q}\right)}\to\infty$ as $q\to\infty$. Following similar discussion as used in Stage 2(i) of Theorem \ref{thm-optsampalloc} for the APCS-B Procedure, given large enough $n_{t_q}$,  $\underset{q\to\infty}{\lim}~\underline{\tilde{s}}_{i,b}^{\left(n_{t_q}\right),b}-\underline{s}_{i,b}^{\left(n_{t_q}\right)}=0$ for $\forall i\ne b$, $\underset{q\to\infty}{\lim}~\left(\underline{\tilde{d}}_{i_0,b}^{\left(n_{t_q}\right),b}\right)^{2}-\left(\underline{d}_{i_0,b}^{\left(n_{t_q}\right)}\right)^{2}>0$, $\underset{q\to\infty}{\lim}~\left(\underline{\tilde{d}}_{i,b}^{\left(n_{t_q}\right),b}\right)^{2}-\left(\underline{d}_{i,b}^{\left(n_{t_q}\right)}\right)^{2}\geq 0$ for $\forall i\ne b,i_0$; meanwhile,  $\underset{q\to\infty}{\lim}~\underline{\tilde{s}}_{i_0,b}^{\left(n_{t_q}\right),i_0}-\underline{s}_{i_0,b}^{\left(n_{t_q}\right)}=0$, $\underset{q\to\infty}{\lim}~\left(\underline{\tilde{d}}_{i_0,b}^{\left(n_{t_q}\right),i_0}\right)^{2}-\left(\underline{d}_{i_0,b}^{\left(n_{t_q}\right)}\right)^{2}=0$. With these conditions,  $\lim_{q\to\infty}\text{mAEOC-B}^{\left(n_{t_q}\right)}-\text{mAEOC-B}^{\left(n_{t_q}\right),b}>0$, $\lim_{q\to\infty}\text{mAEOC-B}^{\left(n_{t_q}\right)}-\text{mAEOC-B}^{\left(n_{t_q}\right),i_0}=0$. It contradicts that $L_{i_0}^{(q)}\to\infty$ as $q\to\infty$. So $\underline{\alpha}_i/\underline{\alpha}_b<\infty$ for $\forall i\ne b$.

\textbf{Stage 2(ii):} Prove that $\lim_{q\to\infty}\underline{d}_{i,b}^{\left(n_{t_q}\right)}\Big{/}\underline{d}_{j,b}^{\left(n_{t_q}\right)}=1$ for $\forall i,j\ne b$. Similar to arguments in analyzing the APCS-B Procedure, for design $i_0\ne b$, $\lim_{q\to\infty}\text{mAEOC-B}^{\left(n_{t_q}\right)}-\text{mAEOC-B}^{\left(n_{t_q}\right),i_0}=\lim_{q\to\infty}\frac{\left(\mu_b-\mu_{i_0}\right)^{2}\cdot\sigma_{i_0}^{2}}{\left(\sigma_{i_0}^{2}+\sigma_b^{2}\alpha_{i_0}^{\left(n_{t_q}\right)}\Big{/}\alpha_b^{\left(n_{t_q}\right)}\right)^{2} n_{t_q}}\cdot\frac{\underline{\nu}_{i_0,b}^{\left(n_{t_q}\right)}+n_{t_q}\underline{\tau}_{i_0}^{\left(n_{t_q}\right)}}{\underline{\nu}_{i_0,b}^{\left(n_{t_q}\right)}-1}\phi_{\underline{\nu}_{i_0,b}^{\left(n_{t_q}\right)}}\left(\underline{d}_{i_0,b}^{\left(n_{t_q}\right)}\right)$. Suppose $\exists\epsilon_0>0$, $\exists i_0,j_0\ne b$, $\underset{q\to\infty}{\lim\sup}~\underline{d}_{i_0,b}^{\left(n_{t_q}\right)}\Big{/}\underline{d}_{j_0,b}^{\left(n_{t_q}\right)}\geq \sqrt{1+\epsilon_0}$. Denote by  $\left\{\underline{\alpha}_i^{\left(n_{t_{q_r}}\right)}\Big{|}L_i^{(r)}\to\infty~\text{as}~r\to\infty,i=1,\dots,M\right\}$ a subsequence of $\left\{\underline{\alpha}_i^{\left(n_{t_q}\right)}\Big{|}L_i^{(q)}\to\infty~\text{as}~q\to\infty,i=1,\dots,M\right\}$ satisfying that $\exists\tilde{\zeta}'>\tilde{\zeta}_0$, $\forall n_{t_{q_r}}>\tilde{\zeta}'$, $\underline{d}_{i_0,b}^{\left(n_{t_{q_r}}\right)}\Big{/}\underline{d}_{j_0,b}^{\left(n_{t_{q_r}}\right)}\geq \sqrt{1+\epsilon_0}$. By similar discussion in analyzing the APCS-B Procedure,  $\lim_{r\to\infty}\frac{\text{mAEOC-B}^{\left(n_{t_{q_r}}\right)}-\text{mAEOC-B}^{\left(n_{t_{q_r}}\right),i_0}}{\text{mAEOC-B}^{\left(n_{t_{q_r}}\right)}-\text{mAEOC-B}^{\left(n_{t_{q_r}}\right),j_0}}=0$. It contradicts that $L_{i_0}^{(q)}\to\infty$ as $q\to\infty$. So $\lim\sup_{q\to\infty}\underline{d}_{i,b}^{\left(n_{t_q}\right)}\Big{/}\underline{d}_{j,b}^{\left(n_{t_q}\right)}\leq 1$ for $\forall i,j\ne b$. Similarly, we can prove that $\lim\inf_{q\to\infty}\underline{d}_{i,b}^{\left(n_{t_q}\right)}\big{/}\underline{d}_{j,b}^{\left(n_{t_q}\right)}\geq 1$ for $\forall i,j\ne b$. Thus, $\lim_{q\to\infty}\underline{d}_{i,b}^{\left(n_{t_q}\right)}\big{/}\underline{d}_{j,b}^{\left(n_{t_q}\right)}=1$ for $\forall i,j\ne b$.

\textbf{Stage 2(iii):} Prove by contradiction that $\underline{\alpha}_i>0$ for $\forall i$. Suppose $\forall i\ne b$, $\underline{\alpha}_{i}^{\left(n_{t_q}\right)}\big{/}\underline{\alpha}_{b}^{\left(n_{t_q}\right)}\to 0$ as $q\to\infty$. By similar discussion in analyzing the APCS-B Procedure,  $\lim_{q\to\infty}\text{mAEOC-B}^{\left(n_{t_q}\right)}-\text{mAEOC-B}^{\left(n_{t_q}\right),b}=0$, $\lim_{q\to\infty}\text{mAEOC-B}^{\left(n_{t_q}\right)}-\text{mAEOC-B}^{\left(n_{t_q}\right),i}>0$, $\forall i\ne b$. It contradicts that $L_{b}^{(q)}\to\infty$ as $q\to\infty$. So $\underline{\alpha}_i/\underline{\alpha}_b>0$ holds for some $i\ne b$. Furthermore, suppose $\exists i_0\ne b$, $\underline{\alpha}_{i_0}^{\left(n_{t_q}\right)}\big{/}\underline{\alpha}_{b}^{\left(n_{t_q}\right)}\to 0$ as $q\to\infty$. By similar discussion in analyzing the APCS-B Procedure, $\lim_{q\to\infty}\underline{\tau}_{i_0}^{\left(n_{t_q}\right)}\Big{/}\underline{\tau}_{j_0}^{\left(n_{t_q}\right)}=0$, which contradicts that $\lim_{q\to\infty}\underline{\tau}_{i}^{\left(n_{t_q}\right)}\Big{/}\underline{\tau}_{j}^{\left(n_{t_q}\right)}=1$ for $\forall i,j\ne b$. So $\underline{\alpha}_i/\underline{\alpha}_b>0$ for $\forall i\ne b$. Based on $0<\underline{\alpha}_i/\underline{\alpha}_b<\infty$ for $\forall i\ne b$, it holds that $\underline{\alpha}_i>0$ for $\forall i$.

\textbf{Stage 2(iv):} Prove that $\underset{q\to\infty}{\lim}\frac{\text{mAEOC-B}^{\left(n_{t_q}\right)}-\text{mAEOC-B}^{\left(n_{t_q}\right),i}}{\text{mAEOC-B}^{\left(n_{t_q}\right)}-\text{mAEOC-B}^{\left(n_{t_q}\right),j}}$\\$=1$, $\forall i,j$. Suppose $\underset{q\to\infty}{\lim\sup}\frac{\text{mAEOC-B}^{\left(n_{t_q}\right)}-\text{mAEOC-B}^{\left(n_{t_q}\right),i_0}}{\text{mAEOC-B}^{\left(n_{t_q}\right)}-\text{mAEOC-B}^{\left(n_{t_q}\right),j_0}}$\\$>1$ for some $i_0,j_0$. We can find a subsequence $\left\{\underline{\alpha}_i^{\left(n_{t_{q_r}}\right)}\Big{|}L_i^{(r)}\to\infty~\text{as}~r\to\infty,i=1,\dots,M\right\}$ satisfying $\underset{r\to\infty}{\lim}\frac{\text{mAEOC-B}^{\left(n_{t_{q_r}}\right)}-\text{mAEOC-B}^{\left(n_{t_{q_r}}\right),i_0}}{\text{mAEOC-B}^{\left(n_{t_{q_r}}\right)}-\text{mAEOC-B}^{\left(n_{t_{q_r}}\right),j_0}}>1$. It contradicts that $L_{j_0}^{(r)}\to\infty$ as $r\to\infty$. So it holds that $\underset{q\to\infty}{\lim\sup}\frac{\text{mAEOC-B}^{\left(n_{t_q}\right)}-\text{mAEOC-B}^{\left(n_{t_q}\right),i}}{\text{mAEOC-B}^{\left(n_{t_q}\right)}-\text{mAEOC-B}^{\left(n_{t_q}\right),j}}\leq 1$ for $\forall i,j$. By similar discussion as used above, we can prove that $\underset{q\to\infty}{\lim\inf}\frac{\text{mAEOC-B}^{\left(n_{t_q}\right)}-\text{mAEOC-B}^{\left(n_{t_q}\right),i}}{\text{mAEOC-B}^{\left(n_{t_q}\right)}-\text{mAEOC-B}^{\left(n_{t_q}\right),j}}\geq 1$ for $\forall i,j$. Thus, $\underset{q\to\infty}{\lim}\frac{\text{mAEOC-B}^{\left(n_{t_q}\right)}-\text{mAEOC-B}^{\left(n_{t_q}\right),i}}{\text{mAEOC-B}^{\left(n_{t_q}\right)}-\text{mAEOC-B}^{\left(n_{t_q}\right),j}}=1$ for $\forall i,j$.

\textbf{Stage 2(v):} Prove that $\left(\frac{\underline{\alpha}_b}{\sigma_b}\right)^{2}-\sum_{i\ne b}\left(\frac{\underline{\alpha}_i}{\sigma_i}\right)^{2}=0$. By similar discussion in analyzing the APCS-B Procedure,
\begin{align*}
&\frac{\sum_{i\ne b}\left(\frac{\underline{\alpha}_i}{\sigma_i}\right)^{2}}{\left(\frac{\underline{\alpha}_b}{\sigma_b}\right)^{2}}
=\lim_{q\to\infty}\sum_{i\ne b}\frac{\frac{\tilde{\underline{\tau}}_{i}^{\left(n_{t_q}\right),b}-\underline{\tau}_{i}^{\left(n_{t_q}\right)}}{\underline{\alpha}_{b}^{\left(n_{t_q}\right)}+\frac{1}{n_{t_q}}-\underline{\alpha}_{b}^{\left(n_{t_q}\right)}}}{\frac{\tilde{\underline{\tau}}_{i}^{\left(n_{t_q}\right),i}-\underline{\tau}_{i}^{\left(n_{t_q}\right)}}{\underline{\alpha}_{i}^{\left(n_{t_q}\right)}+\frac{1}{n_{t_q}}-\underline{\alpha}_{i}^{\left(n_{t_q}\right)}}}\\
=&\lim_{q\to \infty}\underset{i\ne b}{\sum}\frac{\sqrt{\underline{s}_{i,b}^{\left(n_{t_q}\right)}}\Psi_{\underline{\nu}_{i,b}^{\left(n_{t_q}\right)}}\left(\underline{d}_{i,b}^{\left(n_{t_q}\right)}\right)-\sqrt{\underline{\tilde{s}}_{i,b}^{\left(n_{t_q}\right),b}}\Psi_{\underline{\tilde{\nu}}_{i,b}^{\left(n_{t_q}\right),b}}\left(\tilde{\underline{d}}_{i,b}^{\left(n_{t_q}\right),b}\right)}{\sqrt{\underline{s}_{i,b}^{\left(n_{t_q}\right)}}\Psi_{\underline{\nu}_{i,b}^{\left(n_{t_q}\right)}}\left(\underline{d}_{i,b}^{\left(n_{t_q}\right)}\right)-\sqrt{\underline{\tilde{s}}_{i,b}^{\left(n_{t_q}\right),i}}\Psi_{\underline{\tilde{\nu}}_{i,b}^{\left(n_{t_q}\right),i}}\left(\tilde{\underline{d}}_{i,b}^{\left(n_{t_q}\right),i}\right)}\\
=&\lim_{q\to \infty}\frac{\text{mAEOC-B}^{\left(n_{t_q}\right)}-\text{mAEOC-B}^{\left(n_{t_q}\right),b}}{\text{mAEOC-B}^{(n_{t_q})}-\text{mAEOC-B}^{\left(n_{t_q}\right),j}}
=1,~\forall j\ne b.
\end{align*}
That is, $\left(\frac{\underline{\alpha}_b}{\sigma_b}\right)^{2}-\sum_{i\ne b}\left(\frac{\underline{\alpha}_i}{\sigma_i}\right)^{2}=0$. Thus, $\left\{\underline{\alpha}_i\big{|}i=1,\dots,M\right\}$ satisfies the optimality conditions (\ref{ocba-nor}). According to Lemma \ref{lem3}, the sample allocation $\left\{\underline{\alpha}_i^{(n)}\Big{|}i=1,\dots,M\right\}$ from the mAEOC-B Procedure satisfies (\ref{thm-opteqs1}) and (\ref{thm-opteqs2}).	

\textbf{Stage 3:} Prove that $\left\{\alpha_i^{(n)}\Big{|}i=1,\dots,M\right\}$ from the AEOC-B Procedure satisfies (\ref{lem3-eq1}). By similar discussion in analyzing the APCS-B Procedure, $\exists\epsilon'$, $\exists\bar{\zeta}'$, $\forall n>\bar{\zeta}'$, in the case of (\ref{thm2-apcsb-d-ub}),  
\begin{align*}
    &\text{AEOC-B}^{\left(n\right)}-\text{AEOC-B}^{\left(n\right),i}\\
    \geq&\frac{\mu_b-\mu_i}{\sqrt{\tau_i^{(n)}+\epsilon'}}\Psi_{\nu_{i,b}^{\left(n\right)}}\left(\sqrt{n\left(\tau_i^{(n)}+\epsilon'\right)}\right)-\frac{\mu_b-\mu_i}{\sqrt{\tilde{\tau}_i^{(n),i}+\epsilon'}}\cdot\Psi_{\tilde{\nu}_{i,b}^{\left(n\right),i}}\left(\sqrt{n\left(\tilde{\tau}_i^{(n),i}+\epsilon'\right)}\right)\triangleq\underline{\vartheta}_{i}^{(n)}
\end{align*}
for $\forall i\ne b$, 
\begin{align*}
    &\text{AEOC-B}^{\left(n\right)}-\text{AEOC-B}^{\left(n\right),b}\\
    \geq&\sum_{i\ne b}\left[\frac{\mu_b-\mu_i}{\sqrt{\tau_i^{(n)}+\epsilon'}}\Psi_{\nu_{i,b}^{\left(n\right)}}\left(\sqrt{n\left(\tau_i^{(n)}+\epsilon'\right)}\right)-\frac{\mu_b-\mu_i}{\sqrt{\tilde{\tau}_i^{(n),b}+\epsilon'}}\Psi_{\tilde{\nu}_{i,b}^{\left(n\right),b}}\left(\sqrt{n\left(\tilde{\tau}_i^{(n),b}+\epsilon'\right)}\right)\right]\triangleq\underline{\vartheta}_{b}^{(n)}.
\end{align*}
In the case of (\ref{thm2-apcsb-d-lb}), we can derive that
\begin{align*}
    &\text{AEOC-B}^{\left(n\right)}-\text{AEOC-B}^{\left(n\right),i}\\
    \leq&\frac{\mu_b-\mu_i}{\sqrt{\tau_i^{(n)}-\epsilon'}}\Psi_{\nu_{i,b}^{\left(n\right)}}\left(\sqrt{n\left(\tau_i^{(n)}-\epsilon'\right)}\right)-\frac{\mu_b-\mu_i}{\sqrt{\tilde{\tau}_i^{(n),i}-\epsilon'}}\Psi_{\tilde{\nu}_{i,b}^{\left(n\right),i}}\left(\sqrt{n\left(\tilde{\tau}_i^{(n),i}-\epsilon'\right)}\right)\triangleq\overline{\vartheta}_{i}^{(n)}
\end{align*}
for $\forall i\ne b$; meanwhile, 
\begin{align*}
    &\text{AEOC-B}^{\left(n\right)}-\text{AEOC-B}^{\left(n\right),b}\\
    \leq&\sum_{i\ne b}\left[\frac{\mu_b-\mu_i}{\sqrt{\tau_i^{(n)}-\epsilon'}}\Psi_{\nu_{i,b}^{\left(n\right)}}\left(\sqrt{n\left(\tau_i^{(n)}-\epsilon'\right)}\right)-\frac{\mu_b-\mu_i}{\sqrt{\tilde{\tau}_i^{(n),b}-\epsilon'}}\Psi_{\tilde{\nu}_{i,b}^{\left(n\right),b}}\left(\sqrt{n\left(\tilde{\tau}_i^{(n),b}-\epsilon'\right)}\right)\right]\triangleq\overline{\vartheta}_{b}^{(n)}.
\end{align*}
Now we consider a hybrid procedure: When $n\leq\bar{\zeta}'$, the sample allocation is generated based on the AEOC-B Procedure. When $n>\bar{\zeta}'$, the sample allocation is manipulated by a slightly adjusted version of mAEOC-B Procedure (called the \~{m}AEOC-B Procedure). In the \~{m}AEOC-B Procedure, one always replaces $\text{mAEOC-B}^{\left(n\right)}-\text{mAEOC-B}^{\left(n\right),i}$ by $\underline{\vartheta}_{i}^{(n)}$ to evaluate non-best design $i\ne b$, and replaces $\text{mAEOC-B}^{\left(n\right)}-\text{mAEOC-B}^{\left(n\right),b}$ by $\underline{\vartheta}_{b}^{(n)}$ to evaluate design $b$ in the case of (\ref{thm2-apcsb-d-ub}); meanwhile, one always replaces $\text{mAEOC-B}^{\left(n\right)}-\text{mAEOC-B}^{\left(n\right),i}$ by $\overline{\vartheta}_{i}^{(n)}$ to evaluate non-best design $i\ne b$, and replaces $\text{mAEOC-B}^{\left(n\right)}-\text{mAEOC-B}^{\left(n\right),b}$ by $\overline{\vartheta}_{b}^{(n)}$ to evaluate design $b$ in the case of (\ref{thm2-apcsb-d-lb}). Following similar arguments in analyzing the APCS-B Procedure, the sample allocation from the AEOC-B Procedure is lower-bounded by that from the \~{m}AEOC-B Procedure in the case of (\ref{thm2-apcsb-d-ub}), and is upper-bounded by that from the \~{m}AEOC-B Procedure in the case of (\ref{thm2-apcsb-d-lb}). For the \~{m}AEOC-B Procedure, $\lim_{n\to\infty}\left|\tau_i^{(n)}-\tau_j^{(n)}\right|\leq 2\epsilon'$ for $\forall i,j\ne b$, $\lim_{n\to\infty}\left(\alpha_{b}^{\left(n\right)}\big{/}\sigma_{b}\right)^{2}-\sum_{i\ne b}\left(\alpha_{i}^{\left(n\right)}\big{/}\sigma_{i}\right)^{2}\leq c\cdot\epsilon'$ for some finite $c>0$. Then, as $\epsilon'\to 0$,  $\lim_{n\to\infty}\tau_i^{(n)}-\tau_j^{(n)}=0$ for $\forall i,j\ne b$, $\lim_{n\to\infty}\left(\alpha_{b}^{\left(n\right)}\big{/}\sigma_{b}\right)^{2}-\sum_{i\ne b}\left(\alpha_{i}^{\left(n\right)}\big{/}\sigma_{i}\right)^{2}=0$ for the AEOC-B Procedure. Thus, $\left\{\alpha_i^{(n)}\Big{|}i=1,\dots,M\right\}$ from the AEOC-B Procedure satisfies (\ref{thm-opteqs1}) and (\ref{thm-opteqs2}).

The proof of Theorem \ref{thm-optsampalloc} for the APCS-S Procedure has the same idea as above and are omitted for brevity.
\hfill $\square$

\end{document}